\documentclass[11pt]{article}

\usepackage[final]{acl}

\usepackage{times}
\usepackage{latexsym}
\usepackage{times}
\usepackage{latexsym}
\usepackage{amsmath}
\usepackage{booktabs}
\usepackage{multirow}
\usepackage{multicol}
\usepackage{amssymb}
\usepackage[T1]{fontenc}
\usepackage[utf8]{inputenc}
\usepackage{microtype}
\usepackage{inconsolata}
\usepackage{graphicx}
\usepackage{enumitem}

\usepackage[table]{xcolor}
\usepackage[T1]{fontenc}

\usepackage[utf8]{inputenc}

\usepackage{microtype}

\usepackage{inconsolata}

\usepackage{graphicx}
\usepackage{amsmath,amssymb}
\usepackage{mathtools}
\usepackage{amsthm}
\usepackage{multirow}
\usepackage{setspace}
\usepackage{tcolorbox}
\usepackage{latexsym}
\usepackage{subfigure}
\usepackage[table]{xcolor}
\usepackage[capitalize,noabbrev]{cleveref}
\usepackage{hyperref}
\tcbuselibrary{theorems}
\usepackage{bbding}
\usepackage{pifont}
\usepackage{algorithm}
\usepackage{algpseudocode}
\theoremstyle{plain}

\theoremstyle{definition}

\theoremstyle{remark}

\newcommand{\cmark}{\ding{51}}%
\newcommand{\xmark}{\ding{55}}%

\usepackage[textsize=tiny]{todonotes}
\definecolor{tablecolor}{hsb}{0.7, 0.1, 0.98} 
\definecolor{mygo}{RGB}{197, 224, 180}
\newcommand{\lc}{\cellcolor{mygo}}
\newcommand{\cc}{\cellcolor{tablecolor}}
\usepackage{listings}
\usepackage{xcolor}

\newtcolorbox{DefinitionBox}{
  colback=blue!5,
  colframe=blue!80,
  boxrule=0.5pt,
  arc=2pt,
  left=2pt,
  right=2pt,
  top=2pt,
  bottom=2pt,
}

\newtcolorbox{CorollaryBox}{
  colback=gray!5,
  colframe=gray!80,
  boxrule=0.5pt,
  arc=2pt,
  left=2pt,
  right=2pt,
  top=2pt,
  bottom=2pt,
}

\title{Learning Query-Specific Rubrics from Human Preferences \\ for DeepResearch Report Generation}

\author{
\textbf{Changze Lv\textsuperscript{1,2}},
\textbf{Jie Zhou\textsuperscript{1}},
\textbf{Wentao Zhao\textsuperscript{1}},
\textbf{Jingwen Xu\textsuperscript{2}},
\textbf{Shihan Dou\textsuperscript{2}},
\\
\textbf{Zisu Huang\textsuperscript{2}},
\textbf{Muzhao Tian\textsuperscript{2}},
\textbf{Xiaohua Wang \textsuperscript{2}},
\textbf{Zhengkang Guo \textsuperscript{2}},
\textbf{Yang Liu\textsuperscript{3}}, 
\\
\textbf{Pluto Zhou\textsuperscript{1}},
\textbf{Tao Gui\textsuperscript{2}},
\textbf{Le Tian\textsuperscript{1}},
\textbf{Xiao Zhou\textsuperscript{1}},
\textbf{Xiaoqing Zheng\textsuperscript{2}},
\textbf{Xuanjing Huang\textsuperscript{2}},
\textbf{Jie Zhou\textsuperscript{1}}
\\
\textsuperscript{1}Tencent
\textsuperscript{2}Fudan University
\textsuperscript{3}Tsinghua University
\\
\textbf{Correspondence:} Xiaoqing Zheng (zhengxq@fudan.edu.cn)
}

\begin{document}
\maketitle

\begin{abstract}
Nowadays, developing reliable DeepResearch-style long-form report generation remains challenging, as training and evaluation lack verifiable reward signals.
Accordingly, rubric-based evaluation has become a common practice.
However, existing approaches either rely on coarse, pre-defined rubrics that lack sufficient granularity or depend on manually constructed query-specific rubrics that are costly and difficult to scale.
In this paper, we propose a pipeline to train preference-grounded query-specific rubric generators tailored for DeepResearch report generation.
We first construct a dataset of DeepResearch-style queries annotated with human preferences over paired reports, and train rubric generators via reinforcement learning with a hybrid reward combining preference consistency, format validity, and LLM-based rubric evaluation.
We evaluate the resulting rubric generators in two stages.
First, on a held-out human-preference test set, the learned rubrics discriminate preferred from rejected reports more effectively than generic, prompted, or SFT-trained rubric alternatives.
Second, when used as reward signals to train DeepResearch systems, our rubric generators yield substantial performance gains under both a simple single-agent ReAct framework and a complex multi-agent workflow on the DeepResearch Bench.
Our code and model are available at \url{https://huggingface.co/fdu-lcz/rubric_generator}.
\end{abstract}

\section{Introduction}
Large language models (LLMs)~\cite{achiam2023gpt,guo2025deepseek,yang2025qwen3} have recently enabled DeepResearch systems~\cite{qwen2025deepresearch,google2025gemini,openai2025deepresearch,claude2025research} that can synthesize evidence from large-scale document collections and produce long-form analytical reports for complex, open-ended queries.
Unlike short-form DeepResearch tasks like BrowseComp~\cite{wei2025browsecomp,zhou2025browsecomp}, GAIA~\cite{mialon2023gaia}, and HLE \cite{phan2025humanity}, report generation requires models to reason, retrieve, and integrate over diverse sources and multiple turns, while presenting results in a coherent and well-structured manner.

\begin{figure}[t]
\centering
\includegraphics[width=0.80 \linewidth]{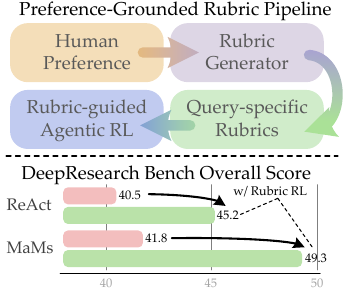}
\vspace{-3mm}
\caption{
Rubric generator trained with human preferences provides query-specific reward signals for rubric-guided agentic RL, improving DeepResearch Bench overall scores under both ReAct and our built multi-agent (MaMs) workflows.
}
\label{fig:intro_performance}
\vspace{-5mm}
\end{figure}

However, developing reliable DeepResearch-style long-form report generators remains fundamentally challenging.
This difficulty stems largely from the lack of verifiable reward signals for both training and evaluation.
Therefore, rubric-based evaluation~\cite{gunjal2025rubrics,huang2025reinforcement,viswanathan2025checklists} has become a practical alternative.
Prior work has explored pre-defined generic rubrics~\cite{que2024hellobench,hashemi2024llm,shao2024assisting} or LLM-generated query-specific rubrics~\cite{xie2025auto,Du2025DeepResearchBA} to provide structured feedback for report generation tasks.
However, those methods suffer from two limitations:
First, pre-defined rubrics are necessarily generic and lack the granularity needed to distinguish subtle quality differences across diverse research queries.
Second, LLM-generated rubrics are typically not grounded in human preferences or domain expertise, making them prone to misalignment with how humans actually compare and judge research reports.
As a result, rubrics produced by both approaches can be noisy or incomplete, yielding weak supervision signals, reward hacking, and inefficient learning dynamics.
In contrast, existing high-quality rubrics~\cite{sharma2025researchrubrics,dou2026cl,dou2026cllife} are typically authored by human experts; this process requires substantial domain expertise and effort for each query, making it difficult to scale to large and diverse training corpora.

In this paper, we argue that one of the most direct supervision signals for assessing report quality is human preference~\cite{dai2023safe,zheng2023secrets,wang2024secrets,liu2024rahf} over candidate reports.
Furthermore, pairwise preferences are easier to collect than manually written query-specific rubrics, but they are too coarse to provide fine-grained training feedback for long-form report generation directly.
This motivates a middle path: using human preferences to learn a reusable rubric generator that converts coarse pairwise judgments into query-specific, informative reward signals.

Therefore, we propose a pipeline to effectively train preference-grounded query-specific rubric generators tailored for DeepResearch report generation.
We construct a preference dataset of over $5,000$ DeepResearch-style queries, each paired with two candidate reports and annotated with human preference judgments.
Then we train the rubric generator using Group Relative Policy Optimization (GRPO)~\cite{shao2024deepseekmath} with a hybrid reward including preference consistency, format compliance, and LLM-based rubric-quality feedback.
Once trained, the rubric generator will be integrated into the training of DeepResearch systems.
For each input query, it automatically produces query-level rubrics that are used to evaluate rollout samples, assigning fine-grained reward scores that guide the optimization process.

Empirically, we demonstrate that our proposed rubric generators deliver more preference-discriminative supervision signals than pre-defined or LLM-generated alternatives.
Furthermore, when employed as training signals for DeepResearch agents, these generated rubrics significantly improve the performance on the downstream benchmarks.
More broadly, our work points to a reusable and scalable pathway for providing fine-grained reward signals in settings where verifiable rewards are scarce: instead of relying on costly expert-written rubrics, systems can learn to derive query-specific supervision from more economical human preference annotations.

To sum up, our contributions are as follows:
\begin{itemize}
\vspace{-3mm}
\setlength{\itemsep}{0pt}
\setlength{\parsep}{0pt}
\setlength{\parskip}{0pt}
\item We construct a large-scale expert preference dataset for DeepResearch reports, using pairwise judgments as a scalable supervision source for query-specific evaluation.
\item We train query-specific rubric generators with GRPO, using a hybrid reward that combines preference consistency, format validity, and LLM-based rubric quality feedback.
\item We demonstrate that the learned rubrics improve both held-out preference discrimination and downstream DeepResearch training in both single- and multi-agent workflows.
\end{itemize}

\section{Related Work}
\subsection{DeepResearch Agent}

\textbf{Short-Form Question Answering.}\quad
In this setting, DeepResearch agents primarily target retrieval-based short-form question answering tasks.
Benchmarks such as GAIA~\cite{mialon2023gaia,russell2025gaia}, BrowseComp~\cite{wei2025browsecomp,zhou2025browsecomp}, and HLE~\cite{phan2025humanity} provide verifiable targets, enabling agent training via Reinforcement Learning with Verifiable Rewards (RLVR)~\cite{jin2025searchr,liu2025webexplorer}.
Several systems leverage this paradigm to enhance search and reasoning capabilities.
For instance, Search-R1~\cite{jin2025searchr} and WebExplorer~\cite{liu2025webexplorer} adopt GRPO to improve retrieval effectiveness in short-form QA tasks with explicit correctness signals.
In contrast, WebThinker~\cite{li2025webthinker} employs Direct Preference Optimization (DPO)~\cite{rafailov2023direct} to equip LLMs with DeepResearch capabilities without relying on verifiable rewards.
Meanwhile, Tongyi DeepResearch~\cite{team2025tongyi} is specifically designed to support long-horizon information-seeking behaviors.

\textbf{Long-Form Report Generation.}\quad
By contrast, long-form report generation requires agents to synthesize evidence from large, heterogeneous document collections and to produce coherent, well-structured reports that address complex, open-ended queries. Beyond retrieving isolated facts, agents must perform multi-step reasoning, reconcile conflicting evidence, and organize information at the document level.
Because evaluating long-form outputs is inherently difficult due to a lack of reference answers, benchmarks in this regime (e.g., DeepResearch Bench~\cite{Du2025DeepResearchBA,li2026deepresearch}, ResearchQA~\cite{Yifei2025ResearchQAES}, and ResearchRubrics~\cite{sharma2025researchrubrics}) commonly use LLM-as-a-Judge applied to human-annotated general or query-specific rubrics.
Recent studies have focused on designing end-to-end workflows for report synthesis.
For example, WebWeaver~\cite{Li2025WebWeaverSW} develops a dual-agent framework that emulates collaborative human research processes.
Dr Tulu~\cite{shao2025dr} and AgentCPM-Report~\cite{li2026agentcpm} are fully open-source DeepResearch agents for long-form tasks.

\subsection{Rubrics for Reward Modeling}\label{sec:related2}
Prior work has explored both fixed rubrics~\cite{hashemi2024llm,que2024hellobench,shao2024assisting} and query-specific rubrics~\cite{shao2025dr,xie2025auto,viswanathan2025checklists} for providing evaluative feedback on outputs produced by agent systems.
More recently, several studies have further incorporated rubrics as reward signals within RL frameworks.
RLCF~\cite{viswanathan2025checklists} and RaR~\cite{gunjal2025rubrics} use checklists (i.e., rubrics) as rewards for downstream task training, yielding fine-grained and multi-criteria supervision.
However, they do not evaluate this approach in non-verifiable-reward settings such as DeepResearch.
CARMO~\cite{gupta2025carmo} proposes inference-time rubric generation for general reward modeling.
For preference-aware rubric generation, P-GenRM~\cite{zhang2026pgenrm} transforms human preferences into evaluation chains that derive adaptive personas and scoring rubrics, while P-Check~\cite{seo2026pcheck} provides dynamic rubric generation for personalized reward modeling.
Web-Shepherd~\cite{chae2025web} introduces a benchmark with preference pairs and annotated checklists for evaluating process reward models.

\begin{figure*}[h]
\centering
\includegraphics[width=1.0\linewidth]{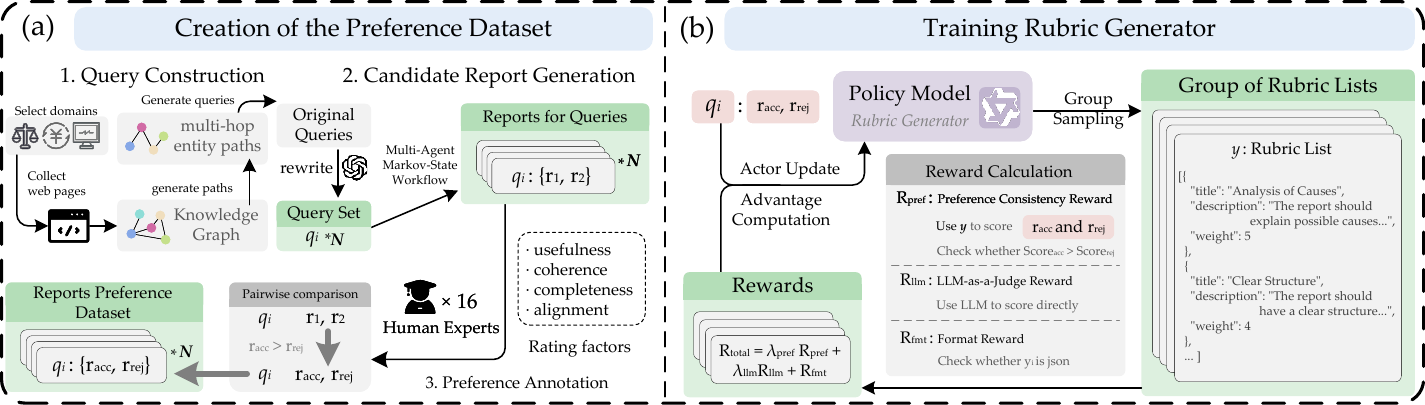}
\caption{Overview of our method.
\textbf{(a)} We construct diverse reporting queries and generate candidate reports. Human experts provide pairwise preference annotations based on usefulness, coherence, completeness, and alignment.
\textbf{(b)} Given a query and its preferred and rejected reports, we train a rubric generator via GRPO to produce weighted evaluation rubrics, with rewards based on preference consistency, LLM-as-a-judge scores, and format validity.
}
\label{fig:main_fig}
\vspace{-4mm}
\end{figure*}

\section{Method}

\subsection{Motivations}

Although expert annotators can provide high-quality evaluation rubrics, manually designing query-specific rubrics at scale is fundamentally impractical.
While such approaches are effective, their reliance on intensive expert effort incurs substantial annotation costs and limits their applicability to large-scale DeepResearch training.
Our objective is not to eliminate human cost entirely, but to transform a recurring annotation burden into a reusable source of supervision.
Instead of asking experts to manually write a structured set of criteria and weights for every new query, we collect pairwise preferences over candidate reports and amortize this one-time supervision into a rubric generator that can be applied to new queries.


\subsection{Creation of the Preference Dataset}\label{sec:pre_dataset}

\begin{figure}[t]
\centering
\includegraphics[width=1.0\linewidth]{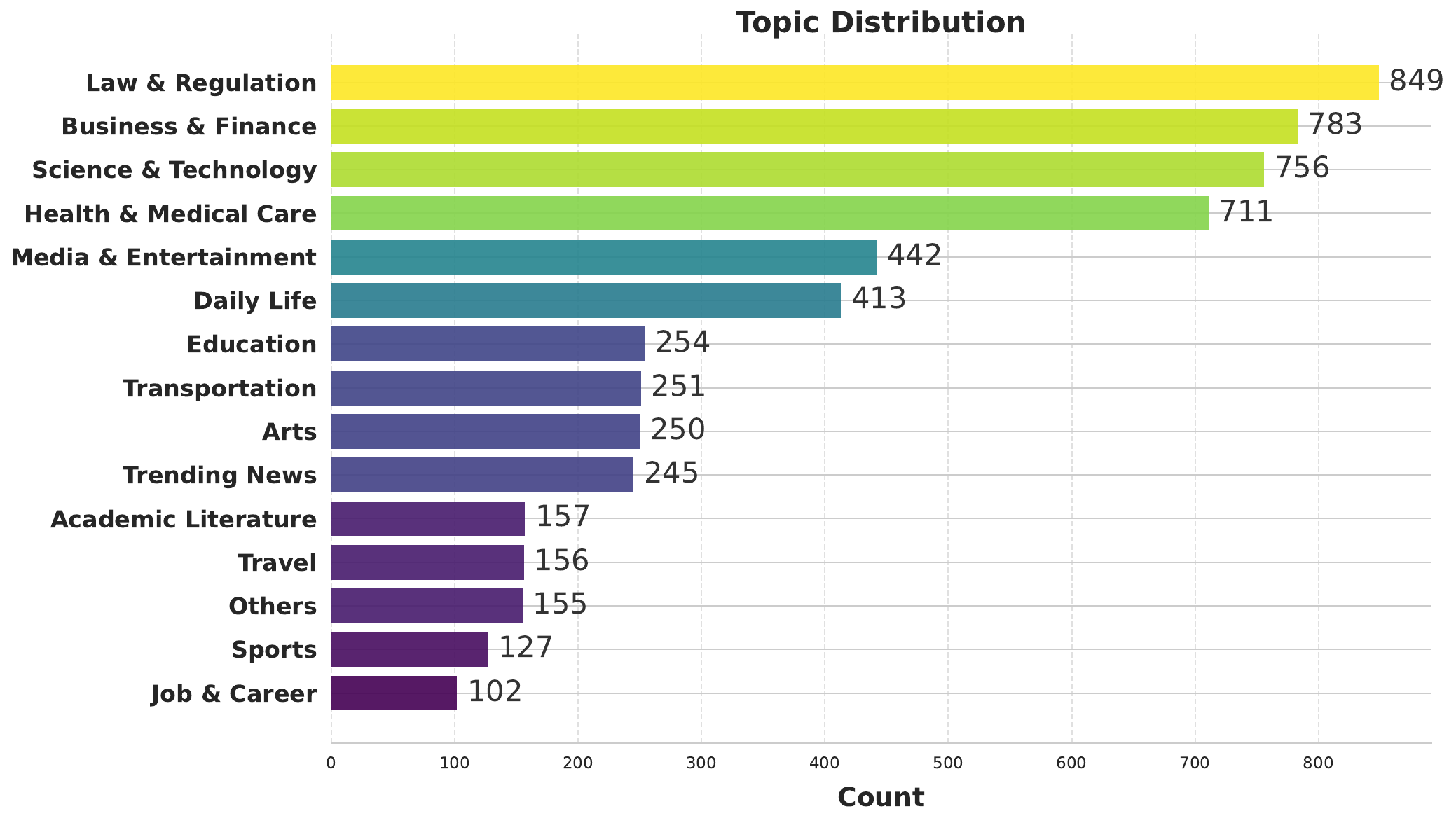}
\caption{Topic distribution of our created human preference dataset for DeepResearch reports.}
\label{fig:topic}
\vspace{-5mm}
\end{figure}

\textbf{Stage 1: Query Construction}

We begin by constructing a diverse set of research-oriented queries that reflect realistic information needs in DeepResearch scenarios.
Each query $q$ is formulated as an open-ended research prompt that requires multi-step reasoning, evidence synthesis, and structured long-form reporting, rather than short factual answers.

Our original queries are automatically generated from a knowledge graph constructed over entities from diverse domains.
The construction follows an iterative process: we first seed the graph with entities from diverse domains, retrieve web pages related to each entity, extract additional entities and relational links from those pages, attach source information to each node, and then sample multi-hop subgraphs for query synthesis.
By leveraging the relational structure of the graph, we sample multi-hop entity paths and prompt an LLM to synthesize corresponding natural-language questions.
This ensures each query is grounded in entity relations while requiring reasoning across multiple facts, making it suitable for evaluating deep research and synthesis.

We additionally rewrite queries with GPT-5~\cite{gpt5} to diversify their phrasing and naturalness, aligning them with realistic user questioning styles and naturally inducing variation in report quality.
Formally, we denote the dataset $\mathcal{Q}$ with $N$ constructed queries as $\mathcal{Q} = \{q_i\}_{i=1}^{N}$, where each $q_i$ serves as the conditioning input for generating subsequent candidate reports.
The case study of rewriting queries and detailed categories is shown in Appendix~\ref{app:query}.

Figure~\ref{fig:topic} summarizes the topic distribution of our query dataset, which spans diverse domains commonly encountered in DeepResearch scenarios, such as \emph{Law}, \emph{Business}, \emph{Science}, and \emph{Health}, along with a long tail of other types.

\textbf{Stage 2: Candidate Report Generation via Multi-Agent Markov State Framework}

Given a fixed query $q \in \mathcal{Q}$, we generate multiple candidate reports by varying hyperparameters across multiple LLMs, including DeepSeek V3.1~\cite{liu2024deepseek3.1} and Tongyi-DeepResearch~\cite{team2025tongyi}, all of which have been trained on agentic data and support tool calling.

To address the challenges arising from long-context dependencies in ReAct-style reasoning~\cite{yao2022react} and the multi-step nature of automated DeepResearch, we draw inspiration from prior work~\cite{Li2025WebWeaverSW,yu2025memagent,chen2025iterresearch} and use a Multi-Agent Markov-State (MaMs) workflow, described in detail in Appendix~\ref{app:algo}.

Using this workflow, report candidates are generated independently and without access to any human annotations.
Before being submitted for human annotation, all candidates undergo a filtering process involving both human reviewers and an auxiliary LLM-based verifier.
This step removes reports with evident factual errors, disorganized or inconsistent citations, or content that exhibits superficial aggregation without coherent reasoning, ultimately retaining only the two highest-quality reports for annotation.
We retain challenging high-quality pairs rather than pairing a strong report with an obviously weak one, because trivially separable pairs provide less useful supervision for learning fine-grained rubrics.

\textbf{Stage 3: Preference Annotation by Human Experts}

For human annotation, we recruit $16$ human experts, each holding at least a master's degree and capable of critically reading and evaluating long-form research reports.
The experts perform pairwise comparisons between candidate reports generated for the same query.
Given a query $q$ and two candidate reports $r_a, r_b \in \mathcal{R}(q)$, annotators are asked to select the report they prefer overall, considering factors such as usefulness, coherence, completeness, and alignment with the information need expressed in $q$, details in Appendix~\ref{app:annotation}.
Each comparison is independently annotated by at least 3 experts, with annotators rotated across queries to reduce individual bias.
The final label is determined by majority vote, resulting in a preferred report $r_{\text{acc}}$ and a less preferred report $r_{\text{rej}}$, forming a preference triple $(q, r_{\text{acc}}, r_{\text{rej}})$.
Aggregating all annotated comparisons yields the final human preference dataset $\mathcal{D} = \{(q, r_{\text{acc}}, r_{\text{rej}})\}$, which is used as supervision for modeling and evaluating preference-aligned report generation.
By relying on expert relative judgments rather than absolute ratings, the dataset captures fine-grained human preferences that are difficult to express with generic or LLM-generated evaluation metrics.

\subsection{Training Rubric Generators with Hybrid Rewards}

To generate evaluation rubrics that better reflect human preferences over reports, we train the rubric generator using GRPO.
Given a query $q$, the policy model $\pi_\theta$ samples a group of rubric candidates $\{y_1, y_2, \ldots, y_G\}$, where each candidate $y_i$ specifies a structured set of evaluation criteria.
Following the rubric specification proposed by~\citet{gunjal2025rubrics}, each rubric item is represented with three key fields: \texttt{title}, \texttt{description}, and an associated importance \texttt{weight}, forming a weighted list of assessment dimensions (e.g., in JSON format).
To robustly guide learning, we design a hybrid reward function $R_{\text{total}}$ that integrates three complementary signals: a preference consistency reward, a format reward, and an LLM-as-a-Judge quality reward.
Formally, the overall training signal is computed as a weighted combination of the above components:
\begin{equation}\label{equ:hybrid_reward}
R_{\text{total}} = \lambda_{\text{pref}} R_{\text{pref}} + \lambda_{\text{llm}} R_{\text{llm}} + R_{\text{fmt}}
\end{equation}

\textbf{Preference Consistency Reward ($R_{\text{pref}}$).}\quad
An effective rubric must be \emph{discriminative}, i.e., capable of reflecting human preferences when applied to real reports.
To this end, we leverage the preference dataset $\mathcal{D}$ that consists of triplets $(q, r_{\text{acc}}, r_{\text{rej}})$, where $r_{\text{acc}}$ is preferred by human annotators over $r_{\text{rej}}$ for the same query $q$.
Given a generated rubric $y$, we score a report $r$ by computing the weighted average of item-level ratings:
\begin{equation}\label{equ:weighted_score}
S(r \mid y) = \frac{\sum_{k=1}^{K} w_k \cdot v_k}{\sum_{k=1}^{K} w_k},
\end{equation}
where $w_k$ denotes the weight of the $k$-th rubric item and $v_k$ is the corresponding conformity score assigned by a judge LLM.
Each $v_k$ is rated on a 1-10 Likert scale~\cite{zheng2023judging,kim2024prometheus} and linearly normalized to the range $[0,1]$ for aggregation.
The preference consistency reward is determined by whether the rubric correctly ranks the human-preferred report above the rejected one:
\begin{equation}
R_{\text{pref}}(y) =
\begin{cases}
+1, & \text{if } S(r_{\text{acc}} \mid y) > S(r_{\text{rej}} \mid y); \\
-1, & \text{otherwise}.
\end{cases}
\vspace{-2mm}
\end{equation}

\textbf{Format Reward ($R_{\text{fmt}}$).}\quad
Secondly, we enforce structural validity as a hard constraint.
Since the downstream evaluation pipeline requires machine-parsable rubric representations, each generated candidate is checked for compliance with the required JSON schema (including mandatory fields such as title, description, and weight).
Candidates that fail this check receive a penalty of $-1$, while structurally valid rubrics incur no additional reward.

\textbf{LLM-as-a-Judge Reward ($R_{\text{llm}}$).}\quad
To assess the intrinsic quality of the generated rubrics, we further adopt an LLM-as-a-Judge mechanism that serves as a semantic meta-evaluator.
Rather than relying on pre-defined rules, the judge evaluates a rubric $y$ in the context of the query $q$, focusing on its logical coherence, coverage comprehensiveness, and the relevance of its evaluation dimensions.
These criteria are synthesized into a scalar quality score $R_{\text{llm}} = \text{Judge}(q, y)$, e.g., scaled to $[0, 4]$.
This reward is used as an auxiliary structural-quality signal; the preference consistency reward remains the only component directly grounded in human comparisons of reports.
The specific prompt can be found in Appendix~\ref{app:prompts_reward_llmjudge}.

\subsection{Rubric-Based GRPO under the Multi-Agent Markov-State Workflow}\label{sec:markov}

After obtaining a trained rubric generator, we leverage it to provide query-specific reward signals for training DeepResearch systems via GRPO.

\begin{figure}[t]
\centering
\includegraphics[width=0.95\linewidth]{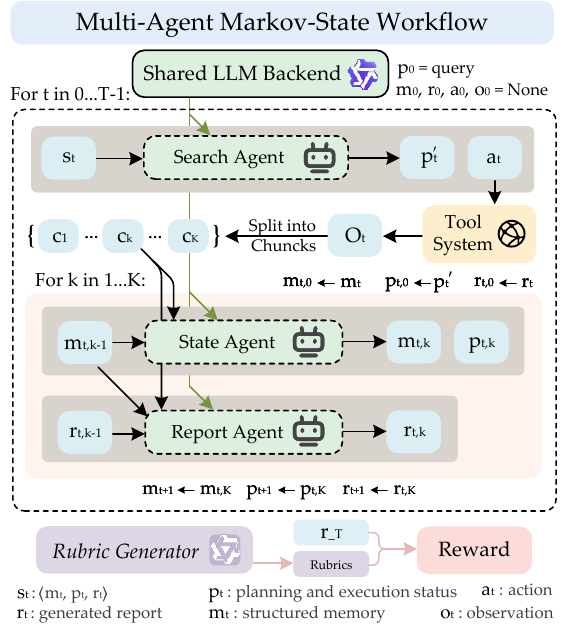}
\caption{Downstream workflow used in rubric-based RL. A shared policy executes search, chunking, state update, and report generation, interacting with external tools. Query-specific rubrics are used to compute rewards of the rollout reports under an individual query.
}
\label{fig:mams}
\vspace{-4mm}
\end{figure}

\textbf{Overview of Multi-Agent Markov-State (MaMs) Workflow.}\quad
We use the MaMs workflow as the DeepResearch system for the downstream RL training.
We do not position MaMs as a standalone new agent paradigm: it shares high-level planning, search, and synthesis patterns with recent DeepResearch systems such as WebThinker~\cite{li2025webthinker} and WebWeaver~\cite{Li2025WebWeaverSW}.
Its role in this paper is to provide a controlled long-context workflow with explicit state variables, shared-policy role prompts, and a stable interface for assigning rubric-based rewards to generated reports.

\begin{table*}[t]
\centering
\caption{
Preference evaluation on the test set of our human preference dataset $\mathcal{D}$.
Acc.\ is preference accuracy/AUC (\%); $d$ is the paired Cohen's $d$ between accepted and rejected reports.
RL refers to GRPO; ``--'' = not applicable.
\textbf{Bold}/\underline{underline} mark the best/second-best per metric within each model.
Human-defined general rubrics (non-model baseline): Acc.\ $=48.78$, $d=0.192$.
}
\vspace{-2mm}
\label{tab:preference_Eval}
\resizebox{0.95\linewidth}{!}{
\begin{tabular}{@{}lcccccccc@{}}
\toprule
& \multicolumn{2}{c}{\textbf{GPT-5}} & \multicolumn{2}{c}{\textbf{Gemini-2.5-Pro}} & \multicolumn{2}{c}{\textbf{Qwen3-14B}} & \multicolumn{2}{c}{\textbf{Qwen3-30B-A3B}} \\
\cmidrule(lr){2-3}\cmidrule(lr){4-5}\cmidrule(lr){6-7}\cmidrule(lr){8-9}
\textbf{Method} & \textbf{Acc.} & \textbf{Cohen's $d$} & \textbf{Acc.} & \textbf{Cohen's $d$} & \textbf{Acc.} & \textbf{Cohen's $d$} & \textbf{Acc.} & \textbf{Cohen's $d$} \\
\midrule
\multicolumn{9}{@{}l}{\emph{Without query-specific rubrics}~(\xmark)} \\
Pointwise Pref.\ Scoring  & $60.28$ & $0.315$ & $57.49$ & $0.297$ & $53.83$ & $0.254$ & $54.01$ & $0.246$ \\
Pairwise Pref.\ Judgment  & $59.93$ & --      & $57.17$ & --      & $53.66$ & --      & $54.53$ & --      \\
\midrule
\multicolumn{9}{@{}l}{\emph{With query-specific rubrics}~({\color{red}\cmark})} \\
Prompted Generation       & $60.80$ & $0.328$ & $59.23$ & $0.302$ & $56.09$ & $0.246$ & $58.54$ & $0.314$ \\
Supervised Fine-tuning    & --      & --      & --      & --      & $59.76$ & $0.260$ & $59.58$ & $0.317$ \\
RL w/ LLM-Judge Reward    & --      & --      & --      & --      & $60.98$ & $0.303$ & $61.50$ & $0.296$ \\
RL w/ Preference Reward   & --      & --      & --      & --      & \underline{$64.63$} & \underline{$0.359$} & \underline{$64.81$} & $\bf0.384$ \\
\cc RL w/ Hybrid Reward   & \cc--   & \cc--   & \cc--   & \cc--   & \cc$\bf65.16$ & \cc$\bf0.366$ & \cc$\bf65.68$ & \cc\underline{$0.376$} \\
\bottomrule
\end{tabular}}
\vspace{-2mm}
\end{table*}

At each turn, MaMs maintains a compact state $s_t=\langle m_t,p_t,r_t\rangle$, where $m_t$, $p_t$, and $r_t$ denote memory, plan, and report, respectively.
A search module selects the next tool action and updates the plan, while state and report modules process retrieved observations chunk by chunk, like MemAgent~\cite{yu2025memagent}, to update memory and draft the report.
All modules are instantiated from the same policy model $\pi_\theta$ and differ only in role-specific prompts and state interfaces.
The detailed state transitions, agent responsibilities, and global algorithm are provided in Appendix~\ref{app:algo}, and the speedup execution is described in Appendix~\ref{app:speed}.

\textbf{Reward Assignment with Weighted Rubrics.}\quad
For each query $q$, the rubric generator produces a list of evaluation rubrics with associated weights, capturing query-specific notions of report quality.
After the system generates a set of candidate reports for $q$, we employ an LLM-as-a-Judge to score each generated report according to these weighted rubrics.
The resulting scalar reward is computed following the same weighted aggregation scheme defined in~\Cref{equ:weighted_score}, and is used to supervise policy optimization.
Please refer to Appendix~\ref{app:prompts} for all prompts.

\section{Experiments}
In this section, we conduct a series of experiments to address the following research questions:

\textbf{RQ1}: Can a rubric generator trained with GRPO effectively capture human preferences over generated reports?

\textbf{RQ2}: Does the rubric generator provide more informative reward signals than general LLMs when used to train DeepResearch agents?

\textbf{RQ3}: How much workflow-level gain does the Markov-state instantiation provide over a conventional ReAct-style framework under the same learned rubric rewards?

\begin{table*}[]
\centering
\caption{Evaluation Results on DeepResearch Bench.
The rubric generator trained by RL refers to the Qwen3-30B-A3B model trained in the previous step using GRPO with hybrid rewards.
Performance of closed-source models is sourced from the official leaderboard, while WebThinker-32B-DPO and DRTulu-Qwen3-8B-RL are obtained from DRTulu. 
\textbf{Bold} font indicates the best performance, whereas \underline{underline} font denotes the second-best performance.
All our models are trained on the queries in the training set of our created dataset $D$.
}
\label{tab:drbench}
\resizebox{0.98\linewidth}{!}{
\begin{tabular}{l|c|c|c|c|c|c|c}
\toprule \hline
\multirow{2}{*}{\textbf{Model}} & \multirow{2}{*}{\textbf{Workflow}} & \multirow{2}{*}{\textbf{Rubric Strategy During RL}} & \multicolumn{5}{c}{\textbf{DeepResearch Bench~\cite{Du2025DeepResearchBA}}} \\ \cline{4-8}
& & & \textbf{Comp.} & \textbf{Depth} & \textbf{Inst.} & \textbf{Read.} & \textbf{Overall}\\ \hline \hline
\multicolumn{8}{c}{\lc\textbf{\emph{Closed-Source DeepResearch}}} \\ \hline
\textbf{OpenAI DeepResearch} & -- & -- &  $46.5$ & $43.7$ & $49.4$ & $47.2$ & $46.5$ \\
\textbf{Claude Research} & -- & -- &  $45.3$ & $42.8$ & $47.6$ & $44.7$ & $45.0$\\
\textbf{Gemini DeepResearch} & -- & -- &  $\bf49.5$ & $\bf49.5$ & \underline{$50.1$} & \underline{$50.0$} & $\bf49.7$\\
\hline
\multicolumn{8}{c}{\lc\textbf{\emph{Open-Source DeepResearch}}} \\ \hline
\textbf{WebThinker-Qwen2.5-32B-DPO} & WebThinker & N/A &  $39.4$ & $35.4$ & $46.0$ & $43.5$ & $40.6$\\
\textbf{DRTulu-Qwen3-8B-RL} & ReAct & Self-Evolving Rubrics &  $41.7$ & $41.8$ & $48.2$ & $41.3$ & $43.4$\\
\textbf{WebWeaver-Qwen3-30B-A3B} & WebWeaver & N/A &  $45.2$ & $45.8$ & $49.6$ & $47.3$ & $46.8$\\
\hline
\multicolumn{8}{c}{\cc\textbf{\emph{Ours}}} \\ \hline
\multirow{3}{*}{\textbf{Qwen3-30B-A3B}} & ReAct & N/A & $33.8$ & $29.9$ & $39.1$ & $40.0$ & $35.0$ \\
& ReAct & Rubric Generator Trained by \textbf{RL} & $38.5$ & $37.9$ & $46.0$ & $43.4$ & $41.0$ \\
& MaMs & Rubric Generator Trained by \textbf{RL} & $44.8$ & $46.4$ & $49.0$ & $47.5$ & $47.2$ \\
\hline
\multirow{7}{*}{\textbf{Tongyi-DeepResearch}} & ReAct & N/A &  $39.5$ & $34.4$ & $46.2$ & $44.3$ & $40.5$\\
&  ReAct & Rubric Generator Trained by \textbf{RL} &$43.4$ & $42.5$ & $48.8$ & $48.0$ & $45.2$ \\ 
\cline{2-8}
& MaMs & N/A &  $38.9$ & $38.5$ & $47.3$ & $44.5$ & $41.8$\\
& MaMs & Human-defined General Rubrics &  $40.5$ & $39.8$ & $48.2$ & $45.4$ & $42.9$\\
& MaMs & GPT-5 Generated Rubrics &  $41.1$ & $39.7$ & $48.5$ & $46.5$ & $43.4$\\
& MaMs & Rubric Generator Trained by \textbf{SFT} &  $40.5$ & $40.9$ & $48.2$ & $45.7$ & $43.4$\\
& \cc MaMs &  \cc Rubric Generator Trained by \textbf{RL} &  \cc \underline{$48.3$} &  \cc \underline{$48.1$} &  \cc $\bf50.7$ &  \cc $\bf50.8$ &  \cc \underline{$49.3$} \\
\hline \bottomrule
\end{tabular}
}
\vspace{-3mm}
\end{table*}

\subsection{Experimental Settings}

\textbf{Datasets.}\quad
For \textbf{RQ1}, we partition $\mathcal{D}$ into train/val/test splits with an $8{:}1{:}1$ ratio at the query level in a topic-balanced manner, so that every topic appears in all splits.
For \textbf{RQ2} and \textbf{RQ3}, we additionally evaluate on DeepResearch Bench~\cite{Du2025DeepResearchBA} ($50$ Chinese and $50$ English research queries).

\textbf{Implementation Details.}\quad
Rubric generators are trained on $8$ NVIDIA H20 GPUs and DeepResearch agents on $32$ H20s; an additional $192$-GPU vLLM~\cite{kwon2023efficient} cluster serves Qwen3-235B-A22B for rubric scoring and LLM-as-a-Judge inference during RL.
Full hyperparameters and infrastructure details are provided in Appendix~\ref{app:imple}.

\textbf{Metrics.}\quad
On $\mathcal{D}$, we report \emph{preference accuracy} (equivalent to pairwise AUC) for ranking correctness and \emph{paired Cohen's $d$}~\cite{diener2010cohen} for the magnitude and stability of preference separation.
Formal definitions are deferred to Appendix~\ref{app:metrics}.
On DeepResearch Bench, we follow the official protocol and report \emph{comprehensiveness}, \emph{depth}, \emph{instruction following}, and \emph{readability}, each judged by LLM judges using the official prompts.
These metrics evaluate the downstream utility of rubrics as scoring and training signals rather than directly auditing every generated rubric item.

\textbf{Baselines.}\quad
On the preference test set, we compare against two families: (i) non-rubric scoring methods (\emph{Human-defined General Rubrics}~\cite{yao2025rigorous}, \emph{Pointwise} and \emph{Pairwise Preference Judgment}), and (ii) rubric-based methods (\emph{Prompted Generation}, \emph{SFT} with GPT-5-generated targets, and GRPO with different reward weightings in~\Cref{equ:hybrid_reward}).
On DeepResearch Bench, closed-source baselines are taken from the official leaderboard, and we further compare ReAct with our MaMs workflow under the same learned rubrics.
Due to infrastructure constraints, we report DRTulu-Qwen3-8B-RL as a reference point rather than a strictly controlled same-backbone comparison.
Full baseline descriptions and protocol details are provided in Appendix~\ref{app:baselines}.

\subsection{Evaluation on Human Preferences}\label{sec:exp_human_pref}

We report the results on the test set of the human preference dataset in \Cref{tab:preference_Eval}, which directly addresses \textbf{RQ1}.
Several key observations can be drawn:

\textbf{(1) Methods based on query-specific rubrics outperform those relying on human-defined general rubrics on our held-out preference test set.}
As shown in the first block of \Cref{tab:preference_Eval}, general rubrics yield near-random preference accuracy and a small effect size, whereas generated, query-conditioned rubrics substantially improve both preference accuracy and paired Cohen's~$d$.
This suggests that incorporating query-specific evaluation criteria is crucial for providing more informative reward signals in downstream training.

\textbf{(2) Directly applying strong LLMs (e.g., GPT-5) to generate rubrics, or performing supervised fine-tuning on such rubrics, does not sufficiently capture fine-grained human preferences.}
Although these approaches achieve gains in preference accuracy, their paired Cohen's~$d$ remains relatively small and does not exhibit a clear separation between accepted and rejected reports, indicating limited alignment with human preference margins.

\textbf{(3) Reinforcement learning with preference-based rewards leads to a pronounced improvement in paired Cohen's~$d$ across both Qwen backbones.}
The increasing Cohen's~$d$ reflects a growing score margin between accepted and rejected reports, suggesting that the model becomes progressively better at discriminating reports preferred by humans.
Moreover, RL with hybrid rewards achieves the best preference accuracy, while preference-only RL remains close and attains the largest effect size for Qwen3-30B-A3B.
This pattern indicates that the human preference reward is the primary signal, with the format and LLM-based quality rewards serving as auxiliary constraints rather than replacing human-grounded supervision.

\subsection{Results on DeepResearch Bench}

Through \Cref{tab:drbench}, we can answer \textbf{RQ2} and \textbf{RQ3}.
First, comparing different rubric strategies under the same Tongyi-DeepResearch backbone, we observe that the rubric generator trained with RL consistently achieves the best performance across all evaluation dimensions.
It outperforms human-defined general rubrics, GPT5-generated rubrics, and the SFT-trained generator, indicating that the proposed rubric generator provides more informative reward signals for training DeepResearch agents under this benchmark protocol.
This validates \textbf{RQ2}, demonstrating that learning rubrics through reinforcement learning yields more effective supervision compared to static or purely supervised alternatives.

We note that Tongyi-DeepResearch exhibits stronger tool-calling and execution abilities than Qwen3-30B-A3B, while being less specialized in report generation than WebWeaver agents, as reflected in the table.
This makes it a suitable backbone for our study, as it allows us to better examine the effectiveness of rubric learning under a capable but not report-specialized DeepResearch system.

Second, the ReAct and MaMs rows estimate the additional gain from this Markov-state workflow under the same learned rubric rewards.
MaMs improves over the corresponding ReAct-style runs in our implementation, suggesting that explicit state variables and chunk-level state updates help apply rubric-based RL under long-context constraints.
We interpret this as an implementation-level gain for the downstream instantiation, rather than as evidence that MaMs is a fundamentally new agent architecture.
In particular, Tongyi-DeepResearch, equipped with the MaMs workflow and an RL-trained rubric generator, achieves the strongest overall performance among the open-source methods reported in \Cref{tab:drbench}, with clear gains in comprehensiveness, instruction following, readability, and overall score.

\subsection{Analysis on Tool Calling}\label{app:tool}

During the training of DeepResearch agents under both the ReAct and MaMs frameworks, the maximum number of interaction turns is set to 10, with up to 5 tool invocations permitted per turn. 
Statistics of interaction turns and tool invocations per sample for our trained DeepResearch systems on DeepResearch Bench are reported in~\Cref{tab:tool}.
As expected, the Tongyi-DeepResearch model, trained on agentic data, exhibits substantially stronger tool-use and interaction capabilities than the vanilla Qwen3-30B-A3B (Instruct) model.
Moreover, DeepResearch systems trained under the MaMs workflow demonstrate superior interaction performance compared to those following the ReAct (search-then-generate) paradigm.

\begin{table}[t]
\centering
\small
\caption{Interaction turns and tool calls per sample of our DeepResearch systems on DeepResearch Bench.}
\label{tab:tool}
\vspace{-3mm}
\begin{tabular}{l l cc}
\toprule
Workflow & Model & Tool Calls & Turns \\
\midrule
ReAct & Qwen3-30B-A3B & 6.05 & 2.21 \\
ReAct & Tongyi-DR & 8.10 & 3.02 \\
\midrule
MaMs & Qwen3-30B-A3B & 19.70 & 7.74 \\
MaMs & Tongyi-DR & \textbf{39.23} & \textbf{9.40} \\
\bottomrule
\end{tabular}
\vspace{-4mm}
\end{table}

\subsection{Analysis on Human Annotation}

To validate the reliability of $\mathcal{D}$, which contains a mixture of Chinese and English queries, we measure inter-annotator agreement (IAA) over all $5{,}651$ pairwise comparisons, each independently labeled by $3$ experts drawn from our pool of $16$.
We obtain a Fleiss' $\kappa$~\cite{fleiss1971measuring} of $0.428$ (bootstrap $95\%$ CI $[0.412, 0.444]$) and an average pairwise Cohen's $\kappa$ of $0.412$, both in the ``moderate agreement'' range~\cite{landis1977measurement}, with $71.4\%$ raw pairwise agreement and $57.8\%$ unanimous ($3{:}0$) labels.
By topic, agreement ranges from more verifiable domains (\emph{Law \& Regulation} $0.503$, \emph{Science \& Technology} $0.476$) to inherently subjective ones (\emph{Arts} $0.348$, \emph{Daily Life} $0.371$).
Because experts themselves disagree on roughly $29\%$ of the pairs, this human pairwise agreement of $71.4\%$ acts as a practical upper bound for any preference-driven model trained on these labels.
Our rubric generator reaches $65.68\%$ accuracy on the held-out test set (\Cref{tab:preference_Eval}), closing most of the gap to this human-agreement ceiling and indicating that the remaining error is largely driven by genuinely ambiguous samples rather than by limited model capacity.

\section{Conclusion}

In this work, we address a core challenge in DeepResearch report generation: obtaining scalable, preference-informed supervision without explicit golden signals.
Instead of relying on predefined or human-annotated rubrics, we learn query-specific rubric generators from human preferences.
Combined with format constraints and LLM-based rubric evaluation under GRPO, our method produces discriminative and adaptable rubrics that provide stronger training and evaluation signals in our experiments.
More broadly, our results highlight learning evaluative criteria as a promising direction for preference-aligned training in complex tasks.
Future work is discussed in Appendix~\ref{app:limit}.

\section*{Limitations}
Despite the improvements demonstrated by our approach, several limitations remain. 
First, the current preference formulation relies solely on pairwise comparisons due to cost limitation, which may not fully capture the complexity of human judgments.
Leveraging richer preference signals, such as rankings or graded scores, could enable more fine-grained learning of human preferences.
Second, the assessment of qualities such as novelty, creativity, factuality, and reasoning depth remains challenging.
Current evaluations are partially subjective, and future work could combine more sophisticated LLM assessments with targeted human feedback to improve both reliability and consistency.


\bibliography{acl}

@article{achiam2023gpt,
  title={Gpt-4 technical report},
  author={Achiam, Josh and Adler, Steven and Agarwal, Sandhini and Ahmad, Lama and Akkaya, Ilge and Aleman, Florencia Leoni and Almeida, Diogo and Altenschmidt, Janko and Altman, Sam and Anadkat, Shyamal and others},
  journal={arXiv preprint arXiv:2303.08774},
  year={2023}
}

@article{guo2025deepseek,
  title={Deepseek-r1: Incentivizing reasoning capability in llms via reinforcement learning},
  author={Guo, Daya and Yang, Dejian and Zhang, Haowei and Song, Junxiao and Zhang, Ruoyu and Xu, Runxin and Zhu, Qihao and Ma, Shirong and Wang, Peiyi and Bi, Xiao and others},
  journal={arXiv preprint arXiv:2501.12948},
  year={2025}
}

@article{yang2025qwen3,
  title={Qwen3 technical report},
  author={Yang, An and Li, Anfeng and Yang, Baosong and Zhang, Beichen and Hui, Binyuan and Zheng, Bo and Yu, Bowen and Gao, Chang and Huang, Chengen and Lv, Chenxu and others},
  journal={arXiv preprint arXiv:2505.09388},
  year={2025}
}

@article{team2025tongyi,
  title={Tongyi deepresearch technical report},
  author={Tongyi, DeepResearch and Li, Baixuan and Zhang, Bo and Zhang, Dingchu and Huang, Fei and Li, Guangyu and Chen, Guoxin and Yin, Huifeng and Wu, Jialong and Zhou, Jingren and others},
  journal={arXiv preprint arXiv:2510.24701},
  year={2025}
}

@article{li2025webthinker,
  title={Webthinker: Empowering large reasoning models with deep research capability},
  author={Li, Xiaoxi and Jin, Jiajie and Dong, Guanting and Qian, Hongjin and Wu, Yongkang and Wen, Ji-Rong and Zhu, Yutao and Dou, Zhicheng},
  journal={arXiv preprint arXiv:2504.21776},
  year={2025}
}

@article{wei2025browsecomp,
  title={Browsecomp: A simple yet challenging benchmark for browsing agents},
  author={Wei, Jason and Sun, Zhiqing and Papay, Spencer and McKinney, Scott and Han, Jeffrey and Fulford, Isa and Chung, Hyung Won and Passos, Alex Tachard and Fedus, William and Glaese, Amelia},
  journal={arXiv preprint arXiv:2504.12516},
  year={2025}
}

@article{zhou2025browsecomp,
  title={Browsecomp-zh: Benchmarking web browsing ability of large language models in chinese},
  author={Zhou, Peilin and Leon, Bruce and Ying, Xiang and Zhang, Can and Shao, Yifan and Ye, Qichen and Chong, Dading and Jin, Zhiling and Xie, Chenxuan and Cao, Meng and others},
  journal={arXiv preprint arXiv:2504.19314},
  year={2025}
}

@article{phan2025humanity,
  title={Humanity's last exam},
  author={Phan, Long and Gatti, Alice and Han, Ziwen and Li, Nathaniel and Hu, Josephina and Zhang, Hugh and Zhang, Chen Bo Calvin and Shaaban, Mohamed and Ling, John and Shi, Sean and others},
  journal={arXiv preprint arXiv:2501.14249},
  year={2025}
}

@article{gunjal2025rubrics,
  title={Rubrics as rewards: Reinforcement learning beyond verifiable domains},
  author={Gunjal, Anisha and Wang, Anthony and Lau, Elaine and Nath, Vaskar and He, Yunzhong and Liu, Bing and Hendryx, Sean},
  journal={arXiv preprint arXiv:2507.17746},
  year={2025}
}

@article{huang2025reinforcement,
  title={Reinforcement learning with rubric anchors},
  author={Huang, Zenan and Zhuang, Yihong and Lu, Guoshan and Qin, Zeyu and Xu, Haokai and Zhao, Tianyu and Peng, Ru and Hu, Jiaqi and Shen, Zhanming and Hu, Xiaomeng and others},
  journal={arXiv preprint arXiv:2508.12790},
  year={2025}
}

@article{viswanathan2025checklists,
  title={Checklists are better than reward models for aligning language models},
  author={Viswanathan, Vijay and Sun, Yanchao and Ma, Shuang and Kong, Xiang and Cao, Meng and Neubig, Graham and Wu, Tongshuang},
  journal={arXiv preprint arXiv:2507.18624},
  year={2025}
}

@article{xie2025auto,
  title={Auto-Rubric: Learning to Extract Generalizable Criteria for Reward Modeling},
  author={Xie, Lipeng and Huang, Sen and Zhang, Zhuo and Zou, Anni and Zhai, Yunpeng and Ren, Dingchao and Zhang, Kezun and Hu, Haoyuan and Liu, Boyin and Chen, Haoran and others},
  journal={arXiv preprint arXiv:2510.17314},
  year={2025}
}

@article{shao2025dr,
  title={Dr tulu: Reinforcement learning with evolving rubrics for deep research},
  author={Shao, Rulin and Asai, Akari and Shen, Shannon Zejiang and Ivison, Hamish and Kishore, Varsha and Zhuo, Jingming and Zhao, Xinran and Park, Molly and Finlayson, Samuel G and Sontag, David and others},
  journal={arXiv preprint arXiv:2511.19399},
  year={2025}
}

@inproceedings{hashemi2024llm,
  title={LLM-Rubric: A Multidimensional, Calibrated Approach to Automated Evaluation of Natural Language Texts},
  author={Hashemi, Helia and Eisner, Jason and Rosset, Corby and Van Durme, Benjamin and Kedzie, Chris},
  booktitle={Proceedings of the 62nd Annual Meeting of the Association for Computational Linguistics (Volume 1: Long Papers)},
  pages={13806--13834},
  year={2024}
}

@article{que2024hellobench,
  title={Hellobench: Evaluating long text generation capabilities of large language models},
  author={Que, Haoran and Duan, Feiyu and He, Liqun and Mou, Yutao and Zhou, Wangchunshu and Liu, Jiaheng and Rong, Wenge and Wang, Zekun Moore and Yang, Jian and Zhang, Ge and others},
  journal={arXiv preprint arXiv:2409.16191},
  year={2024}
}

@inproceedings{shao2024assisting,
  title={Assisting in writing wikipedia-like articles from scratch with large language models},
  author={Shao, Yijia and Jiang, Yucheng and Kanell, Theodore and Xu, Peter and Khattab, Omar and Lam, Monica},
  booktitle={Proceedings of the 2024 Conference of the North American Chapter of the Association for Computational Linguistics: Human Language Technologies (Volume 1: Long Papers)},
  pages={6252--6278},
  year={2024}
}

@article{zheng2023judging,
  title={Judging llm-as-a-judge with mt-bench and chatbot arena},
  author={Zheng, Lianmin and Chiang, Wei-Lin and Sheng, Ying and Zhuang, Siyuan and Wu, Zhanghao and Zhuang, Yonghao and Lin, Zi and Li, Zhuohan and Li, Dacheng and Xing, Eric and others},
  journal={Advances in neural information processing systems},
  volume={36},
  pages={46595--46623},
  year={2023}
}

@inproceedings{dai2023safe,
  title={Safe RLHF: Safe Reinforcement Learning from Human Feedback},
  author={Dai, Josef and Pan, Xuehai and Sun, Ruiyang and Ji, Jiaming and Xu, Xinbo and Liu, Mickel and Wang, Yizhou and Yang, Yaodong},
  booktitle={The Twelfth International Conference on Learning Representations},
  year={2023}
}

@article{zheng2023secrets,
  title={Secrets of rlhf in large language models part i: Ppo},
  author={Zheng, Rui and Dou, Shihan and Gao, Songyang and Hua, Yuan and Shen, Wei and Wang, Binghai and Liu, Yan and Jin, Senjie and Liu, Qin and Zhou, Yuhao and others},
  journal={arXiv preprint arXiv:2307.04964},
  year={2023}
}

@article{wang2024secrets,
  title={Secrets of rlhf in large language models part ii: Reward modeling},
  author={Wang, Binghai and Zheng, Rui and Chen, Lu and Liu, Yan and Dou, Shihan and Huang, Caishuang and Shen, Wei and Jin, Senjie and Zhou, Enyu and Shi, Chenyu and others},
  journal={arXiv preprint arXiv:2401.06080},
  year={2024}
}

@article{shao2024deepseekmath,
  title={Deepseekmath: Pushing the limits of mathematical reasoning in open language models},
  author={Shao, Zhihong and Wang, Peiyi and Zhu, Qihao and Xu, Runxin and Song, Junxiao and Bi, Xiao and Zhang, Haowei and Zhang, Mingchuan and Li, YK and Wu, Yang and others},
  journal={arXiv preprint arXiv:2402.03300},
  year={2024}
}

@inproceedings{mialon2023gaia,
  title={Gaia: a benchmark for general ai assistants},
  author={Mialon, Gr{\'e}goire and Fourrier, Cl{\'e}mentine and Wolf, Thomas and LeCun, Yann and Scialom, Thomas},
  booktitle={The Twelfth International Conference on Learning Representations},
  year={2023}
}

@misc{openai2025deepresearch,
  title = {Introducing deep research},
  author = {{OpenAI}},
  year = {2025},
  howpublished = {\url{https://openai.com/index/introducing-deep-research/}},
  note = {Accessed: 2025-02}
}

@misc{qwen2025deepresearch,
  author       = {{Qwen Team}},
  title        = {Qwen {DeepResearch}: When Inspiration Becomes Its Own Reason},
  year         = {2025},
  month        = {November},
  howpublished = {\url{https://qwen.ai/blog?id=qwen-deepresearch}},
  note         = {Accessed: 2025-12-23}
}

@misc{google2025gemini,
  title = {Gemini Deep Research — your personal research assistant},
  author = {{Google}},
  year = {2025},
  howpublished = {\url{https://gemini.google/overview/deep-research/}},
  note = {Accessed: 2025-03}
}

@misc{claude2025research,
  title = {Claude takes research to new places},
  author = {{Anthropic}},
  year = {2025},
  howpublished = {\url{https://www.anthropic.com/news/research}},
  note = {Accessed: 2025-04}
}

@article{russell2025gaia,
  title={Gaia-2: A controllable multi-view generative world model for autonomous driving},
  author={Russell, Lloyd and Hu, Anthony and Bertoni, Lorenzo and Fedoseev, George and Shotton, Jamie and Arani, Elahe and Corrado, Gianluca},
  journal={arXiv preprint arXiv:2503.20523},
  year={2025}
}

@inproceedings{jin2025searchr,
title={Search-R1: Training {LLM}s to Reason and Leverage Search Engines with Reinforcement Learning},
author={Bowen Jin and Hansi Zeng and Zhenrui Yue and Jinsung Yoon and Sercan O Arik and Dong Wang and Hamed Zamani and Jiawei Han},
booktitle={Second Conference on Language Modeling},
year={2025}
}

@article{liu2025webexplorer,
  title={Webexplorer: Explore and evolve for training long-horizon web agents},
  author={Liu, Junteng and Li, Yunji and Zhang, Chi and Li, Jingyang and Chen, Aili and Ji, Ke and Cheng, Weiyu and Wu, Zijia and Du, Chengyu and Xu, Qidi and others},
  journal={arXiv preprint arXiv:2509.06501},
  year={2025}
}

@article{rafailov2023direct,
  title={Direct preference optimization: Your language model is secretly a reward model},
  author={Rafailov, Rafael and Sharma, Archit and Mitchell, Eric and Manning, Christopher D and Ermon, Stefano and Finn, Chelsea},
  journal={Advances in neural information processing systems},
  volume={36},
  pages={53728--53741},
  year={2023}
}

@article{Li2025WebWeaverSW,
  title={WebWeaver: Structuring Web-Scale Evidence with Dynamic Outlines for Open-Ended Deep Research},
  author={Zijian Li and Xin Guan and Bo Zhang and Shen Huang and Houquan Zhou and Shaopeng Lai and Ming Yan and Yong Jiang and Pengjun Xie and Fei Huang and Jun Zhang and Jingren Zhou},
  journal={arXiv preprint arXiv:2509.13312},
  year={2025}
}

@article{Du2025DeepResearchBA,
  title={DeepResearch Bench: A Comprehensive Benchmark for Deep Research Agents},
  author={Mingxuan Du and Benfeng Xu and Chiwei Zhu and Xiaorui Wang and Zhendong Mao},
  journal={ArXiv},
  year={2025},
  volume={abs/2506.11763}
}

@article{Yifei2025ResearchQAES,
  title={ResearchQA: Evaluating Scholarly Question Answering at Scale Across 75 Fields with Survey-Mined Questions and Rubrics},
  author={Li S. Yifei and Allen Chang and Chaitanya Malaviya and Mark Yatskar},
  journal={ArXiv},
  year={2025},
  volume={abs/2509.00496}
}

@article{gpt5,
author ={OpenAI },
title ={GPT-5 System Card},
year = {2025}
}

@article{liu2024deepseek3.1,
  title={Deepseek-v3 technical report},
  author={Liu, Aixin and Feng, Bei and Xue, Bing and Wang, Bingxuan and Wu, Bochao and Lu, Chengda and Zhao, Chenggang and Deng, Chengqi and Zhang, Chenyu and Ruan, Chong and others},
  journal={arXiv preprint arXiv:2412.19437},
  year={2024}
}

@article{chen2025iterresearch,
  title={IterResearch: Rethinking Long-Horizon Agents via Markovian State Reconstruction},
  author={Chen, Guoxin and Qiao, Zile and Chen, Xuanzhong and Yu, Donglei and Xu, Haotian and Zhao, Wayne Xin and Song, Ruihua and Yin, Wenbiao and Yin, Huifeng and Zhang, Liwen and others},
  journal={arXiv preprint arXiv:2511.07327},
  year={2025}
}

@article{yu2025memagent,
  title={MemAgent: Reshaping Long-Context LLM with Multi-Conv RL-based Memory Agent},
  author={Yu, Hongli and Chen, Tinghong and Feng, Jiangtao and Chen, Jiangjie and Dai, Weinan and Yu, Qiying and Zhang, Ya-Qin and Ma, Wei-Ying and Liu, Jingjing and Wang, Mingxuan and others},
  journal={arXiv preprint arXiv:2507.02259},
  year={2025}
}

@inproceedings{yao2022react,
  title={React: Synergizing reasoning and acting in language models},
  author={Yao, Shunyu and Zhao, Jeffrey and Yu, Dian and Du, Nan and Shafran, Izhak and Narasimhan, Karthik R and Cao, Yuan},
  booktitle={The eleventh international conference on learning representations},
  year={2022}
}

@article{sharma2025researchrubrics,
  title={Researchrubrics: A benchmark of prompts and rubrics for evaluating deep research agents},
  author={Sharma, Manasi and Zhang, Chen Bo Calvin and Bandi, Chaithanya and Wang, Clinton and Aich, Ankit and Nghiem, Huy and Rabbani, Tahseen and Htet, Ye and Jang, Brian and Basu, Sumana and others},
  journal={arXiv preprint arXiv:2511.07685},
  year={2025}
}

@article{yao2025rigorous,
  title={A Rigorous Benchmark with Multidimensional Evaluation for Deep Research Agents: From Answers to Reports},
  author={Yao, Yang and Wang, Yixu and Zhang, Yuxuan and Lu, Yi and Gu, Tianle and Li, Lingyu and Zhao, Dingyi and Wu, Keming and Wang, Haozhe and Nie, Ping and others},
  journal={arXiv preprint arXiv:2510.02190},
  year={2025}
}

@inproceedings{kim2024prometheus,
  title={Prometheus 2: An Open Source Language Model Specialized in Evaluating Other Language Models},
  author={Kim, Seungone and Suk, Juyoung and Longpre, Shayne and Lin, Bill Yuchen and Shin, Jamin and Welleck, Sean and Neubig, Graham and Lee, Moontae and Lee, Kyungjae and Seo, Minjoon},
  booktitle={Proceedings of the 2024 Conference on Empirical Methods in Natural Language Processing},
  pages={4334--4353},
  year={2024}
}

@article{diener2010cohen,
  title={Cohen's d},
  author={Diener, Marc J},
  journal={The Corsini encyclopedia of psychology},
  pages={1--1},
  year={2010},
  publisher={Wiley Online Library}
}

@article{zheng2025group,
  title={Group sequence policy optimization},
  author={Zheng, Chujie and Liu, Shixuan and Li, Mingze and Chen, Xiong-Hui and Yu, Bowen and Gao, Chang and Dang, Kai and Liu, Yuqiong and Men, Rui and Yang, An and others},
  journal={arXiv preprint arXiv:2507.18071},
  year={2025}
}

@misc{slime_github,
  author       = {Zilin Zhu and Chengxing Xie and Xin Lv and slime Contributors},
  title        = {slime: An LLM post-training framework for RL Scaling},
  year         = {2025},
  howpublished = {\url{https://github.com/THUDM/slime}},
  note         = {GitHub repository. Corresponding author: Xin Lv},
  urldate      = {2025-06-19}
}

@inproceedings{kwon2023efficient,
  title={Efficient Memory Management for Large Language Model Serving with PagedAttention},
  author={Woosuk Kwon and Zhuohan Li and Siyuan Zhuang and Ying Sheng and Lianmin Zheng and Cody Hao Yu and Joseph E. Gonzalez and Hao Zhang and Ion Stoica},
  booktitle={Proceedings of the ACM SIGOPS 29th Symposium on Operating Systems Principles},
  year={2023}
}

@article{shoeybi2019megatron,
  title={Megatron-lm: Training multi-billion parameter language models using model parallelism},
  author={Shoeybi, Mohammad and Patwary, Mostofa and Puri, Raul and LeGresley, Patrick and Casper, Jared and Catanzaro, Bryan},
  journal={arXiv preprint arXiv:1909.08053},
  year={2019}
}

@article{zheng2024sglang,
  title={Sglang: Efficient execution of structured language model programs},
  author={Zheng, Lianmin and Yin, Liangsheng and Xie, Zhiqiang and Sun, Chuyue Livia and Huang, Jeff and Yu, Cody Hao and Cao, Shiyi and Kozyrakis, Christos and Stoica, Ion and Gonzalez, Joseph E and others},
  journal={Advances in neural information processing systems},
  volume={37},
  pages={62557--62583},
  year={2024}
}

@inproceedings{peng2024yarn,
  title={YaRN: Efficient Context Window Extension of Large Language Models},
  author={Peng, Bowen and Quesnelle, Jeffrey and Fan, Honglu and Shippole, Enrico},
  booktitle={The Twelfth International Conference on Learning Representations},
  year={2024}
}

@inproceedings{liu2024rahf,
  title={Aligning Large Language Models with Human Preferences through Representation Engineering},
    author = "Liu, Wenhao  and
      Wang, Xiaohua  and
      Wu, Muling  and
      Li, Tianlong  and
      Lv, Changze  and
      Ling, Zixuan  and
      JianHao, Zhu  and
      Zhang, Cenyuan  and
      Zheng, Xiaoqing  and
      Huang, Xuanjing",
  booktitle={Association for Computational Linguistics},
  year={2024}
}

@article{li2026deepresearch,
  title={DeepResearch Bench II: Diagnosing Deep Research Agents via Rubrics from Expert Report},
  author={Li, Ruizhe and Du, Mingxuan and Xu, Benfeng and Zhu, Chiwei and Wang, Xiaorui and Mao, Zhendong},
  journal={arXiv preprint arXiv:2601.08536},
  year={2026}
}

@article{dou2026cl,
  title={CL-bench: A Benchmark for Context Learning},
  author={Dou, Shihan and Zhang, Ming and Yin, Zhangyue and Huang, Chenhao and Shen, Yujiong and Wang, Junzhe and Chen, Jiayi and Ni, Yuchen and Ye, Junjie and Zhang, Cheng and others},
  journal={arXiv preprint arXiv:2602.03587},
  year={2026}
}

@article{li2026agentcpm,
  title={AgentCPM-Report: Interleaving Drafting and Deepening for Open-Ended Deep Research},
  author={Li, Yishan and Chen, Wentong and Yan, Yukun and Li, Mingwei and Mei, Sen and Wang, Xiaorong and Liu, Kunpeng and Cong, Xin and Wang, Shuo and Zhang, Zhong and others},
  journal={arXiv preprint arXiv:2602.06540},
  year={2026}
}

@article{dou2026cllife,
  title={CL-bench Life: Can Language Models Learn from Real-Life Context?},
  author={Dou, Shihan and Shen, Yujiong and Huang, Chenhao and Ye, Junjie and Chen, Jiayi and Wang, Junzhe and He, Qianyu and Liu, Shichun and Lv, Changze and Lin, Jiahang and others},
  journal={arXiv preprint arXiv:2604.27043},
  year={2026}
}

@inproceedings{gupta2025carmo,
  title={CARMO: Dynamic Criteria Generation for Context Aware Reward Modelling},
  author={Gupta, Taneesh and Shandilya, Shivam and Zhang, Xuchao and Madhavan, Rahul and Ghosh, Supriyo and Bansal, Chetan and Yao, Huaxiu and Rajmohan, Saravan},
  booktitle={Findings of the Association for Computational Linguistics: ACL 2025},
  pages={2202--2261},
  year={2025}
}

@inproceedings{zhang2026pgenrm,
  title={P-GenRM: Personalized Generative Reward Model with Test-time User-based Scaling},
  author={Zhang, Pinyi and Lin, Ting-En and Wu, Yuchuan and Chen, Jingyang and Wang, Zongqi and Yang, Hua and Ze, Xu and Huang, Fei and Li, Yongbin and Zhang, Kai},
  booktitle={The Fourteenth International Conference on Learning Representations},
  year={2026}
}

@article{seo2026pcheck,
  title={P-Check: Advancing Personalized Reward Model via Learning to Generate Dynamic Checklist},
  author={Seo, Kwangwook and Lee, Dongha},
  journal={arXiv preprint arXiv:2601.02986},
  year={2026}
}

@article{chae2025web,
  title={Web-shepherd: Advancing prms for reinforcing web agents},
  author={Chae, Hyungjoo and Kim, Sunghwan and Cho, Junhee and Kim, Seungone and Moon, Seungjun and Hwangbo, Gyeom and Lim, Dongha and Kim, Minjin and Hwang, Yeonjun and Gwak, Minju and others},
  journal={arXiv preprint arXiv:2505.15277},
  year={2025}
}

@article{fleiss1971measuring,
  title={Measuring nominal scale agreement among many raters.},
  author={Fleiss, Joseph L},
  journal={Psychological bulletin},
  volume={76},
  number={5},
  pages={378},
  year={1971},
  publisher={American Psychological Association}
}

@article{landis1977measurement,
  title={The measurement of observer agreement for categorical data},
  author={Landis, J Richard and Koch, Gary G},
  journal={biometrics},
  pages={159--174},
  year={1977},
  publisher={JSTOR}
}

\newpage
\appendix
\onecolumn

\section{Case Study on Query Rewriting}\label{app:query}

Our created query set covers a broad range of domains relevant to DeepResearch scenarios.
High-frequency categories include \emph{Law \& Regulation}, \emph{Business \& Finance}, \emph{Science \& Technology}, and \emph{Health \& Medical Care}, reflecting common research-oriented information needs that require multi-step reasoning and evidence synthesis.
The dataset also contains a diverse set of medium- and low-frequency topics, such as \emph{Media \& Entertainment}, \emph{Daily Life}, \emph{Education}, \emph{Arts}, and \emph{Trending News}, as well as long-tail domains including \emph{Academic Literature} and \emph{Job \& Career}.
This distribution mirrors realistic usage patterns of DeepResearch systems, supporting the study of human preferences across heterogeneous report-generation tasks.

The following is a case study of the query rewriting:
\begin{DefinitionBox}
\textbf{Original Query:}

In Alice's Adventures in Wonderland, what is the most common eye color corresponding to the real-life cat breed that inspired the Cheshire Cat?

$\\$

\textbf{DeepResearch-style Query:}

Please conduct a study on the Cheshire Cat from Alice's Adventures in Wonderland: identify the most likely real-world cat breed that served as its inspiration, and summarize the breed's most common coat colors along with the typical eye color associated with each coat.

\end{DefinitionBox}

\section{MaMs workflow}\label{app:algo}

\subsection{Global Algorithm}

We show the detailed algorithm description in~\Cref{alg:mams}.

\begin{algorithm}[t]
\caption{Multi-agent Markov-state (MaMs) Workflow}\label{alg:mams}
\begin{algorithmic}[1]
\State \textbf{Input:} User query $q$, maximum iterations $T$
\State \textbf{Initialize:} memory $m_0$, plan $p_0$, report $r_0$
\For{$t = 0$ to $T-1$}
    \State \textbf{Search Agent:} \textcolor{gray}{// high-level controller}
    \State \hspace{1em} Generate action and updated plan
    \State \hspace{1em} $a_t, p'_t = \mathcal{A}_{\text{search}}(q, s_t)$
    \State \hspace{1em} \textcolor{gray}{// $a_t$: search action, $p'_t$: refined plan}
    
    \State Execute action $a_t$ and obtain raw observation $O_t$
    \State Split $O_t$ into semantically coherent chunks $\{c_1, \dots, c_K\}$
    \State \textcolor{gray}{// chunking handles long context and enables incremental processing}
    
    \State \textbf{Initialize chunk-level states:}
    \State \hspace{1em} $m_{t,0} \leftarrow m_t$, $p_{t,0} \leftarrow p'_t$, $r_{t,0} \leftarrow r_t$
    
    \For{$k = 1$ to $K$}
        \State \textbf{State Agent update:} \textcolor{gray}{// incremental memory \& plan update}
        \State \hspace{1em} $m_{t,k}, p_{t,k} = \mathcal{A}_{\text{state}}(q, c_k, m_{t,k-1}, p_{t,k-1})$
        
        \State \textbf{Report Agent update:} \textcolor{gray}{// incremental report generation}
        \State \hspace{1em} $r_{t,k} = \mathcal{A}_{\text{report}}(q, c_k, m_{t,k-1}, r_{t,k-1})$
    \EndFor
    
    \State Update global state after all chunks processed
    \State \hspace{1em} $m_{t+1} \leftarrow m_{t,K}$, $p_{t+1} \leftarrow p_{t,K}$, $r_{t+1} \leftarrow r_{t,K}$
    \State \textcolor{gray}{// $s_{t+1} = \langle m_{t+1}, p_{t+1}, r_{t+1} \rangle$}
    
    \If{termination condition is satisfied} 
        \State \textcolor{gray}{// stop if max turns reached, or plan indicates no further search needed}
        \State \textbf{break}
    \EndIf
\EndFor
\State \textbf{Return:} Final report $r_\text{final}$
\end{algorithmic}
\end{algorithm}

The termination condition is triggered when either (i) the maximum number of interaction turns is reached, and the system is forced to stop and produce a final report, or (ii) the Search Agent determines that no further information acquisition is required according to the current plan.
Therefore, the final report $r_\text{final}$ is conditionally equal to $r_T$.

\subsection{Detailed MaMs Workflow Description}\label{app:mams_detail}

\textbf{State Abstraction and Iterative Transitions.}\quad
We model the deep research process as a sequential decision-making problem over an abstract state space. 
For a user query $q$, the research state at iteration turn $t$ is defined as $s_t = \langle m_t, p_t, r_t \rangle$. 
Here, $m_t$ represents the structured memory, $p_t$ denotes the dynamic execution plan, and $r_t$ is the incrementally evolving report. 
Unlike standard RAG systems that condition on raw retrieved context, our framework operates on this compact abstraction, ensuring scalability across long-horizon workflows.
The transitions follow a hierarchical structure: a high-level search action triggers low-level state processing. 
Formally, $s_{t+1} = \mathcal{T}(s_t, a_t)$, where $\mathcal{T}$ encapsulates the tool execution and the subsequent multi-agent processing pipeline described below.

\textbf{Agent Modules and Chunk-based Process.}\quad
The MaMs workflow consists of $3$ specialized agents with clearly defined responsibilities, shown in~\Cref{fig:main_fig}.

\textbf{\emph{Search Agent:}}
Acting as the high-level controller, the search agent observes the current state $s_t$ and determines the optimal next step.
It generates a search action $a_t$ (e.g., generating \texttt{<tool\_call></tool\_call>}) and refines the global plan: $a_t, p'_{t} = \mathcal{A}_{\text{search}}(q, s_t)$.
This agent is responsible for identifying information gaps in $m_t$ and driving the exploration process.
If sufficient information is gathered, the search loop terminates, and the output is finalized.

\textbf{\emph{State Agent:}}
Upon execution of action $a_t$, the environment returns a raw observation $O_t$ (e.g., long search content).
A critical challenge is that $O_t$ often exceeds the context window limits of the LLM. 
To address this, we follow MemAgent~\cite{yu2025memagent} to implement a chunk-based processing mechanism.
The raw text $O_t$ is segmented into a sequence of smaller chunks $\{c_1, c_2, \dots, c_K\}$ using a text splitter that respects semantic boundaries (e.g., paragraphs).

The state agent processes these chunks sequentially to update the memory and plan while minimizing information loss.
Let $m_{t,0} = m_t$ and $p_{t,0} = p'_t$.
For each chunk $k \in \{1,\dots,K\}$, the agent performs an incremental update:
\begin{equation}
    m_{t,k}, p_{t,k} = \mathcal{A}_{\text{state}}(q, c_k, m_{t,k-1}, p_{t,k-1}).
\end{equation}
The prompt logic explicitly enforces an ``incremental fusion'' strategy: existing knowledge in $m_{t,k-1}$ is preserved, while new facts from $c_k$ are compressed and merged. 
After processing all $K$ chunks, the final state for the next iteration is established as $m_{t+1} = m_{t,K}$ and $p_{t+1} = p_{t,K}$.

\textbf{\emph{Report Agent:}}
The report agent incrementally refines the research report alongside state updates:
\begin{equation}
    r_{t,k} = \mathcal{A}_{\text{report}}(q, c_k, m_{t,k-1}, r_{t,k-1}).
\end{equation}
This design effectively decouples information compression (handled by state agent) from narrative generation (handled by report agent). 
The report agent uses the streaming evidence $c_k$ to draft, correct, and expand sections of the report $r_t$, ensuring global consistency and reducing the hallucination risk associated with generating long reports in a single pass.
Once termination conditions, maximum turns or no further tool calls, are met, the final report $r_\text{final}$ is produced.

Although the system adopts a multi-agent architecture at the functional level, all agents are instantiated from the same LLM.
Specifically, these three agents share a single policy model $\pi_\theta$ and differ only in their role-specific prompts, action spaces, and state interfaces.
As a result, MaMs can also be viewed as a structured single-agent formulation with modularized behaviors, rather than a multi-agent learning system with independently optimized policies.
This design isolates architectural benefits from model heterogeneity, ensuring that observed behaviors arise from agent specialization rather than differences in model capacity.

\section{Prompts Used in MaMs Workflow and LLM-as-a-Judge}\label{app:prompts}
For each prompt, we have both Chinese and English versions, as the question dataset is bilingual.
In this section, we present the English version, while the corresponding Chinese version is included in the supplementary material.
\subsection{Prompts for Generating Query-Specific Rubrics}

\begin{lstlisting}
You are a professional rubric-writing expert. Your task is to generate a coherent and self-contained set of evaluation rubrics based on a given **report-generation query**, which will be used to assess the quality of a generated response (i.e., a report).

Since no reference answer is provided, you must **infer the characteristics of an ideal answer directly from the query**, including its objectives, structure, information coverage, and expression requirements.

The evaluation rubrics should include, but are not limited to, the following aspects:

* Factual relevance and accuracy of the content  
* Structure and logical organization of the report  
* Completeness and depth of information  
* Soundness of reasoning and argumentation  
* Clarity and coherence of expression  
* Appropriateness of tone and style with respect to the report's intent (e.g., summary, analysis, recommendation)

Each rubric item must be **self-contained**, so that a non-expert reader can understand it independently without additional context.  
Each description must begin with one of the following prefixes:

- ``Key Criterion: ...''
- ``Important Criterion: ...''
- ``Optional Criterion: ...''
- ``Error Criterion: ...''

---

### **Input:**
* query: the full text of the report-generation request

### **Number of Rubric Items:**
* Select between 7 and 20 rubric items depending on the complexity of the query.

### **Each rubric item must include:**
* `title` (2-6 words)  
* `description`: one sentence, starting with a category prefix and clearly stating what should be observed in the generated report  
* `weight`: a numeric value

* Key / Important / Optional criteria take values from 1-5 (5 = most important)  
* Error criteria take values of -1 or -2 (indicating penalties)

---

### **Category Definitions:**

* **Key Criterion**: Core facts, structure, or objectives that must be present; missing them makes the answer invalid (weight = 5)
* **Important Criterion**: Critical reasoning, completeness, or clarity that significantly affects quality (weight = 3-4)
* **Optional Criterion**: Stylistic or depth-related enhancements (weight = 1-2)
* **Error Criterion**: Common mistakes or omissions, explicitly indicating ``missing'' or ``incorrect'' elements (weight = -1 or -2)

---

### **Additional Guidelines:**

* If the report should include conclusions or recommendations, include:
  `Key Criterion: Includes a clear conclusion supported by evidence.`
* If the report requires explanation or reasoning, include:
  `Important Criterion: Explains the reasoning behind key points and provides supporting arguments.`
* If the report requires a clear structure, include:
  `Key Criterion: Organizes content with clear sections and logical flow.`
* If the report has a specific tone (e.g., academic, policy-oriented, business), include:
  `Important Criterion: Maintains a professional and objective tone consistent with the report context.`
* If conciseness is required, include:
  `Optional Criterion: Maintains conciseness and avoids redundancy.`

---

### **Output Requirements:**

* Output a JSON array in the format: [{...}, {...}, ...], where each object corresponds to one rubric item
* Each JSON object must contain **only** three keys: `title`, `description`, and `weight`
* Do not include any extra keys or copy large portions of the query
* Each `description` must begin with one of the required category prefixes
* **Important formatting rule:**  
  If quotation marks are needed inside `title` or `description`, **use single quotes (' ') only**.  
  Do NOT use double quotes (" "), as they will break the JSON format.  
  Example: use 'Michelin star' instead of "Michelin star".

---

### **Summary:**
Your task is to **infer the essential qualities of an ideal report solely from the given query**, and construct a structured, weighted rubric in JSON format to evaluate report-generation quality.

Return **only** the requested JSON array. Do not include any additional explanations or text.
\end{lstlisting}

\subsection{Prompts for Scoring a Report by a Single Rubric through LLM-as-a-Judge}
\begin{lstlisting}
You are a precise and impartial scoring model.

Your task: evaluate the degree to which a report aligns with a given single rubric description, based solely on that rubric.

**Input Information**
Query: {query}
Rubric: {rubric}
Report to be scored: {report}

**Scoring Instructions**
- You only need to judge how well the report "matches" the description of the rubric.
- Do not judge whether the rubric represents a positive goal or a negative constraint.
- Do not attempt to reverse or correct the semantic direction of the rubric.
- Do not introduce any additional evaluation criteria.

**Scoring Requirements**
- Output an integer score from 1 to 10:
- 10 = report fully aligns with the rubric description
- 7-9 = largely aligns
- 4-6 = partially aligns
- 1-3 = largely does not align

**Output Format** (strict, single line, no punctuation):
rating: <integer from 1 to 10>
\end{lstlisting}

\subsection{Prompts for LLM-based Judgement of Rubrics}\label{app:prompts_reward_llmjudge}
\begin{lstlisting}
You are an accurate and impartial scoring model (Reward Model). Your task is to evaluate the quality of **rubrics** (evaluation criteria).
A rubric is a set of standards used to assess the quality of model-generated answers. You need to determine whether the given rubric is reasonable, comprehensive, and aligned with the task objective.

Based on the following information, you should assess how well the rubric generated by the policy model (response) matches your criteria.

**[Input Information]**
Question: {question}
Rubric to be evaluated (response): {response}

**[Scoring Requirements]**
You must output three items:

1. **[reward]**: A decimal number ranging from 0.00 to 4.00 (up to two decimal places).

* 4.00 = High quality: clear structure, comprehensive dimensions, rigorous logic, and strong alignment with the question.
* 3.00 = Generally reasonable: covers key dimensions but with minor omissions or less concise expression.
* 2.00 = Partially reasonable: covers some important aspects, but lacks key elements or has notable logical flaws.
* 1.00 = Weakly related: low relevance to the task or serious format issues.
* 0.00 = Completely irrelevant or meaningless: does not meet the evaluation purpose or is empty/garbled.

2. **[confidence]**: Your confidence in the score (0%-100%). A higher value indicates greater certainty.

3. **[reason]**: A brief explanation of the scoring rationale.

**[Important Note]**
You are evaluating the *design quality of the rubric itself*, not the quality of any report or answer.

**[Output Format]** (strictly three lines, no punctuation):

```
reward: <decimal between 0.00 and 4.00>
confidence: <integer percentage between 0% and 100%>
reason: <brief explanation in English>
```
\end{lstlisting}

\subsection{System Prompt of Search Agent in MaMs workflow}

\begin{lstlisting}
You are an intelligent assistant capable of generating high-quality deep research reports. Your goal is to solve complex user problems through multiple cycles of "Plan-Execute-Observe."

### Core Process
1. **Analyze State**: Review the current `<memory>` (information obtained so far) and `<plan>` (current progress).
2. **Develop Strategy**:
   - If information is insufficient or the plan is incomplete -> update the Plan and use tools (e.g., search) to gather information.
   - If information is sufficient and the plan is complete -> organize your thoughts and output the final report.
3. **Output Specifications**:
   - Update the plan table `<plan>...</plan>`: mark completed items and list remaining tasks.
   - Final action: either invoke a tool or output `<answer>...</answer>`.

### Notes
- **Plan**: must be a Markdown list, clearly showing current and upcoming steps.
- **Answer**: generate `<answer>...</answer>` only when you are confident all necessary information has been collected.

**Tool Instructions**: {tool description}
\end{lstlisting}

\subsection{User Prompt of Search Agent in MaMs workflow}
\begin{lstlisting}
<user_input>
{{ query }}
</user_input>

<memory>
{{ memory }}
</memory>

<plan>
{{ plan }}
</plan>

<report>
{{ report }}
</report>

Remaining tool call chances: {{ tool_call_chance }}.
Based on the current state (Memory/Plan) and the completeness of <report>, plan the next action.
If the current <report> is unsatisfactory, continue updating <plan> and use tools to search.
If the <report> is deemed complete, directly output <answer>...</answer> to finish.

Strictly follow the output format:
<plan>Updated execution plan</plan>
<tool_call>Tool invocation details (if any)</tool_call> or <answer>End</answer>
\end{lstlisting}

When the count of the tool calling action meets the threshold (default $10$), then we will change the user prompt as:
\begin{lstlisting}
Tool call chances have been exhausted.
Based on the following information:
<user_input>
{{ query }} 
</user_input>

<memory>
{{ memory }}
</memory>

<plan>
{{ plan }}
</plan>
List your final plan. Do not call tools again.
\end{lstlisting}

\subsection{System Prompt of State Agent in MaMs workflow}
\begin{lstlisting}
You are an information processing expert responsible for maintaining a "long-term memory" database. You are currently in a multi-step process of reading a long text in chunks.

### Task Objective
After reading the current "Observation Fragment," you need to **incrementally merge** newly discovered information into the existing `<memory>`.  
Note: the input `<memory>` contains all previously accumulated key information. **When updating via compression, details are easily lost, so you must take all measures to prevent this.**

### Core Principles
1. **Preserve Old Memory (Most Important)**:
   - Information in the input `<memory>` that is not mentioned in the current fragment **must be retained** in the output.
   - Do not remove information from Memory just because it is absent in the current fragment.
2. **Incremental Integration**:
   - Only add facts, data, or insights that are **new** from the current fragment to Memory.
   - If new information corrects old information, modify it; if it is redundant, ignore it.
3. **Maintain High Density**:
   - Memory should be a "pile of facts," not an article summary.
   - Preserve specific numbers, names, dates, and references. Do not write "a detailed discussion about XX"; instead, write "XX stated that YYY."

### Steps
1. Read the input `<memory>` (old knowledge).  
2. Read the `Tool Output` below (new fragment).  
3. Output a new `<memory>`: it = old memory + new knowledge from the fragment.

### Output Format
Strictly follow this format for use in the next decision step:
<memory>Updated memory integrating old and new information</memory>
\end{lstlisting}
\subsection{User Prompt of State Agent in MaMs workflow}
\begin{lstlisting}
<user_input>
{{ query }}
</user_input>

<memory>
{{ memory }}
</memory>

<plan>
{{ plan }}
</plan>

Please read the following tool output fragment.  
Task: extract key information to update <memory>.

Strictly follow the output format:
<memory>Updated key retrieved information summary</memory>
\end{lstlisting}
\subsection{System Prompt of Report Agent in MaMs workflow}

\begin{lstlisting}
You are a professional structured analysis report writing assistant, responsible for maintaining a <report> that is continuously updated based on user input. Your goal is to incrementally update the existing <report> based on the tool-provided information, **without introducing external information**.

### Workflow
When you receive the user query <user_input>, key information summary <memory>, execution plan <plan>, the current round report <report>, and new information from tool calls, perform the following steps:
1. Analyze the type of new information.
2. Decide whether the new information should be included in the updated report.
3. If it should be included, update the original report:
   - Do not simply append new information; instead, supplement, correct, or replace content while maintaining logical flow.
   - Avoid expanding the scope of content unnecessarily.

### Core Principles
1. Update the report solely based on user-provided information:
   - Do not add external facts, speculative information, fabricated data, or extrapolated scenarios. Do not infer information not present in reality.
   - Do not add uncertainty disclaimers in the report.
2. **Do not simply append new information**:
   - Assess whether new information is relevant; if so, integrate it into the corresponding section. Otherwise, omit it.
   - Structure may be optimized if necessary, but core content must remain stable.
3. Maintain logical consistency:
   - If new information conflicts with the existing report, carefully decide whether to replace the old information based on current knowledge.
   - The report must not contain contradictory statements.

### Report Requirements
1. Output <report> in Markdown format.
2. Ensure <report> has a clear structure, rigorous logic, and high readability.
3. At the end of <report>, list all necessary references or sources (each numbered, with full citation), avoiding duplicates.
4. Citation formatting rules:
   - In the report body, superscript citations may be used, e.g., `<sup>[1]</sup>`.
   - If superscripts are used, the corresponding entry must be included in the "References" section.
   - Superscripts must immediately follow the cited noun or term, not at the beginning of a sentence. Correct examples: ``...the law<sup>[1]</sup> states...'', ``Article 1 of the Civil Code<sup>[4]</sup> stipulates...''.

### Output Format
Strictly follow this format:
<report>Complete report content</report>

\end{lstlisting}

\subsection{User Prompt of Report Agent in MaMs workflow}

\begin{lstlisting}
<user_input>
{{ query }} 
</user_input>

<memory>
{{ memory }}
</memory>

<report>
{{ report }}
</report>
\end{lstlisting}

\subsection{System Prompt in ReAct workflow}

\begin{lstlisting}
You are a deep research expert. You need to use search tools to investigate the question posed by the user and eventually produce a comprehensive and in-depth report. Your research process follows the steps below:

**Research Process**
1. Carefully read and analyze the user's question, considering what information the user needs.
2. Develop a detailed research plan by breaking the user's question into multiple sub-questions. If necessary, further decompose sub-questions until each is simple enough. For each decomposed question, create a search plan.
3. In the same round as planning, perform the first round of tool calls. To increase efficiency, you may generate at most {{max_tool_call_cnt_per_round}} tool calls per message.
4. Enter the "Plan Revision - Search" loop. In each iteration:
   (1) Organize the results returned by the search tools. Consider what information is still missing and whether new leads need to be explored. If needed, revise your search plan and ensure it covers all potential user concerns, adding supplementary searches as necessary.
   (2) Check whether the latest search plan still contains questions that need searching. If so, generate a new round of tool calls, again limited to {{max_tool_call_cnt_per_round}} per message. Then wait for the search results.
   (3) If in step (2) you determine that the search plan is complete and you have enough information to write the report, synthesize the search results into a comprehensive and insightful report through logical inference rather than listing facts. Do not perform further tool calls; the process will automatically end.

**Requirements**
1. Mandatory Tool Calls: While research is ongoing, every assistant message **must include tool_calls**. If a reply contains only text and no tool calls, the task is considered complete.
2. Multi-Round Limit: Complete all research within {{max_turn}} rounds, i.e., a maximum of {{max_turn}} messages.
3. Per-Round Call Limit: Each message may generate at most {{max_tool_call_cnt_per_round}} tool calls.
4. Search Breadth: When developing or revising the search plan, consider all possible directions relevant to the research question and collect as much information and detail as possible.
5. Search Depth: Do not only search for "what it is"; focus on "why" and "how it works." For key phenomena, explore underlying mechanisms or deeper causes. If a current search result mentions a critical concept, technology, or contradiction, prioritize investigating it in the next round rather than switching to a parallel topic. Avoid wasting too many rounds on deep tracing.
6. Report Requirements:
   - Avoid information dumping: The report may be divided into sections, but strictly avoid listing retrieved facts without synthesis. The core value of the report is transforming fragmented information into logically connected, systematic discussion.
   - Logical Completeness: Every main point must have a full argument arc: state the core conclusion -> provide concrete evidence (e.g., data, cases, details) -> explain underlying mechanisms or relevance (i.e., why or what it implies).
   - Substantive Content: Avoid empty adjectives (e.g., "highly effective," "promising"). Use concrete technical parameters, quantitative metrics, regulatory details, or expert opinions from search results.
   - Multi-Dimensional Perspective: For complex issues, analyze from multiple dimensions (e.g., cause analysis, risk assessment, long-term impact, technical path comparison) ensuring each dimension is sufficiently supported.

**Citation Standards**
1. **In-text citations**: Use superscript format in the report body, e.g., "This is an important conclusion<sup>[1]</sup>."
2. **Reference List**: At the end of the report, list all references. Include **full article title and URL**. If the search result does not provide a URL, only include the title. Format:
   [1] Article Title - URL  
   [2] Article Title - URL
3. **Ordering**: Number references in the order of their first appearance in the text.
4. **Deduplication**: If the same source is cited multiple times (even across rounds), merge into a single entry with the same number; do not duplicate.
5. **Source Extraction**: Use the titles from search result summaries formatted as [Title: xxxx] directly as reference names.

**Tool Instructions**: {tool description}
\end{lstlisting}

\subsection{Prompt for Pairwise Preference Judgment}

\begin{lstlisting}
Please act as an impartial judge and evaluate the quality of the responses provided by two AI assistants to the user question displayed below. You should choose the assistant that follows the user's instructions and answers the user's question better. Your evaluation should consider factors such as helpfulness, relevance, accuracy, depth, creativity, and level of detail of their responses. Begin your evaluation by comparing the two responses and provide a short explanation. Avoid any positional biases and ensure that the order in which the responses were presented does not influence your decision. Do not allow the length of the responses to influence your evaluation. Do not favor certain names of the assistants. Be as objective as possible.

User Question: {question}

[The Start of Assistant A's Answer]
{answer_a}
[The End of Assistant A's Answer]

[The Start of Assistant B's Answer]
{answer_b}
[The End of Assistant B's Answer]

Please output your final verdict by strictly following this format: "[[A]]" if Assistant A is better, "[[B]]" if Assistant B is better.
\end{lstlisting}

\subsection{Prompt for Pointwise Preference Scoring}

\begin{lstlisting}
Please act as an impartial judge and evaluate the quality of the response provided by an AI assistant to the user question displayed below. Your evaluation should consider factors such as helpfulness, relevance, accuracy, depth, creativity, and level of detail of the response. You should give a score between 1 and 10, where 1 is the worst and 10 is the best.

User Question: {question}

[The Start of Assistant's Answer]
{answer}
[The End of Assistant's Answer]

Please output your final verdict by strictly following this format: "[[score]]", for example "[[8]]".
\end{lstlisting}

\subsection{Prompts of DeepResearch Bench}

The prompts used to evaluate generated reports on the DeepResearch Bench are directly adopted from the official prompts released on GitHub\footnote{\url{https://github.com/Ayanami0730/deep_research_bench/tree/main/prompt}}.

\section{Implementation Details}\label{app:imple}

Our training code is based on the post-training framework slime\footnote{\url{https://github.com/THUDM/slime}}~\cite{slime_github}, which leverages Megatron~\cite{shoeybi2019megatron} for the training backend and SGlang~\cite{zheng2024sglang} for the inference backend.
Note that there is a crucial update\footnote{\url{https://github.com/THUDM/slime/issues/958}} on the Megatron config of optimizers to this framework, ensuring the correct training of MoE models when reinforcement learning. We show hyperparameters for training rubric generators in \Cref{tab:qwen3_params} and hyperparameters for training DeepResearch Agents based on Tongyi-DeepResearch in \Cref{tab:deep_research_params}.
For evaluation, we follow the official DeepResearch Bench protocol and adopt Gemini-2.5-Pro as the LLM-as-a-Judge.
During reinforcement learning, rubric scoring is conducted using Qwen3-235B-A22B with a temperature of 0.3, top-p of 0.95, and a maximum context length of 131,072 tokens enabled by Yarn RoPE scaling~\cite{peng2024yarn}.
Unless otherwise specified, the weighting coefficients $\lambda_{\text{pref}}$ and $\lambda_{\text{llm}}$ in \Cref{equ:hybrid_reward} are both set to 1.
All policy models, including rubric generators and DeepResearch agents, are trained with a context length of 64k tokens, temperature of 1.0, and top-p of 1.0.
For DeepResearch agents, the maximum number of interaction turns is set to 10, with up to 5 tool invocations allowed per turn.

\begin{table}[htp]
\centering
\small
\caption{Hyperparameters for Training Rubric Generators based on Qwen3-30B-A3B with Hybrid Reward}
\label{tab:qwen3_params}
\begin{tabular}{lc@{\hskip 0.8in}lc}
\toprule
\textbf{Hyperparameter} & \textbf{Value} & \textbf{Hyperparameter} & \textbf{Value} \\
\midrule
\multicolumn{2}{l}{\textit{\textbf{Optimization Config}}} & \multicolumn{2}{l}{\textit{\textbf{GRPO Strategy}}} \\
Optimizer & Adamw & Algorithm & GRPO \\
Learning Rate & $1 \times 10^{-6}$ & Group Size ($G$) & 8 \\
Weight Decay & 0.1 & KL Coefficient & 0.0 \\
Global Batch Size & 256 & Clip Ratio ($\epsilon$) & 0.2 \\
LR Schedule & Constant & Advantage & Group Relative \\
\midrule
\multicolumn{2}{l}{\textit{\textbf{System \& Parallelism}}} & \multicolumn{2}{l}{\textit{\textbf{Generation \& Data}}} \\
Tensor Parallel (TP) & 4 & Max Response Len & 8,192 \\
Expert Parallel (EP) & 8 & Temperature & 1.0 \\
Context Parallel (CP) & 1 & Rollout Batch Size & 32 \\
Max Tokens/GPU & 30,000 &  &  \\
\bottomrule
\end{tabular}
\end{table}

\begin{table}[htp]
\centering
\small
\caption{Hyperparameters for Training DeepResearch Agents}
\label{tab:deep_research_params}
\begin{tabular}{lc@{\hskip 0.8in}lc}
\toprule
\textbf{Hyperparameter} & \textbf{Value} & \textbf{Hyperparameter} & \textbf{Value} \\
\midrule
\multicolumn{2}{l}{\textit{\textbf{Optimization Config}}} & \multicolumn{2}{l}{\textit{\textbf{GRPO Strategy}}} \\
Optimizer & Adam & Algorithm & GRPO \\
Learning Rate & $1 \times 10^{-6}$ & Group Size ($G$) & 8 \\
Weight Decay & 0.1 & KL Coefficient & 0.0 \\
Global Batch Size & 64 & Clip Ratio ($\epsilon$) & 0.2 \\
LR Schedule & Constant & Advantage & Group Relative \\
\midrule
\multicolumn{2}{l}{\textit{\textbf{System \& Parallelism}}} & \multicolumn{2}{l}{\textit{\textbf{Generation \& Data}}} \\
Tensor Parallel (TP) & 4 & Max Response Len & 16,384 \\
Expert Parallel (EP) & 8 & Temperature & 1.0 \\
Context Parallel (CP) & 2 & Rollout Batch Size & 8 \\
Max Tokens/GPU & 6,000 & Observation Window & 24,000 \\
\bottomrule
\end{tabular}
\end{table}



\section{Baselines}\label{app:baselines}

\textbf{Baselines for human preference evaluation.}\quad
We compare against the following methods:
(1) \textbf{Human-defined General Rubrics}, which adopt manually specified evaluation rubrics following the general report rubrics proposed in \citet{yao2025rigorous};
(2) \textbf{Pointwise Preference Scoring}, where the accepted and rejected reports $(r_{\text{acc}}, r_{\text{rej}})$ in each triplet $(q, r_{\text{acc}}, r_{\text{rej}})$ are scored independently by the model, and preference is determined by score comparison;
(3) \textbf{Pairwise Preference Judgment}, where $(r_{\text{acc}}, r_{\text{rej}})$ are jointly provided to the model for direct preference judgment;
(4) \textbf{Generated Rubrics}, which prompt the model to generate query-specific rubrics from $q$, followed by LLM-based evaluation;
(5) \textbf{Supervised Fine-Tuning (SFT)}, which uses GPT-5-generated rubrics as supervision targets;
and (6) \textbf{Reinforcement Learning with Various Rewards}, where GRPO is applied with different reward weight configurations in~\Cref{equ:hybrid_reward}.

\textbf{Baselines for DeepResearch Bench.}\quad
Closed-source baselines are reported directly from the official DeepResearchBench leaderboard\footnote{\url{https://huggingface.co/spaces/muset-ai/DeepResearch-Bench-Leaderboard}}.
Due to infrastructure constraints, we are unable to reproduce the DRTulu system with Qwen3-30B-A3B as the backbone model and the same external search stack; we therefore report the available DRTulu-Qwen3-8B-RL result as a reference point rather than a strictly controlled same-backbone comparison.
We also compare ReAct with our MaMs workflow under the same learned rubrics.
The search tool performs keyword-based retrieval and returns full results, while for the ReAct (search-then-generate) framework we provide a summarized version to prevent output-length issues and ensure effective reasoning.
Because this difference changes the way retrieved evidence is exposed to the generator, ReAct--MaMs comparisons should be interpreted as workflow-level comparisons under practical context-length constraints.
For fairness within our own runs, all results are obtained using the checkpoint with the best validation performance.

\section{Annotation Guidelines for Pairwise Preference Collection}\label{app:annotation}

This appendix presents the full guidelines distributed to our $16$ human annotators for collecting the pairwise preference triples $(q, r_{\text{acc}}, r_{\text{rej}})$ that constitute the dataset $\mathcal{D}$ in~\Cref{sec:pre_dataset}.
We aim to make the protocol explicit and reproducible: from the rendering of the annotation interface, through the operational definitions of the evaluation dimensions, to the bias-mitigation and quality-control measures.
For each guideline, both Chinese and English versions are released together with the dataset; we present the English version here for brevity.

\textbf{Task Overview.}\quad
For every assignment, the annotator is shown a single research-oriented query $q$ and two candidate reports $r_a$ and $r_b$ generated by (anonymized) DeepResearch systems for that same query.
The annotator must read both reports in full and select the report that, \emph{taken as a whole}, better satisfies the information need expressed in $q$.
Annotators do not assign absolute scores and do not write rubrics; the only required output is a binary preference label together with a short free-text justification (1--3 sentences) recording the dominant reason for the choice.
The justification is used solely for downstream auditing and quality control, not for training.

\textbf{Annotator Pool and Calibration.}\quad
All $16$ annotators hold at least a master's degree and have demonstrated, during a screening interview, the ability to critically read long-form analytical text in both Chinese and English.
Before being assigned production batches, each annotator completes a calibration round on $20$ practice pairs whose ``gold'' labels were jointly produced by the authors after extensive discussion.
Annotators must reach $\geq 80\%$ agreement with the gold labels and pass a follow-up debrief with the lead annotator before they are admitted to production.
This calibration step is intended to align the annotators' interpretations of the four evaluation dimensions described below, rather than to enforce a single ``correct'' aesthetic preference.

\textbf{Annotation Interface.}\quad
Each annotation page presents the query at the top, followed by two reports rendered side-by-side and labelled neutrally as ``Report~1'' and ``Report~2''.
The labels are independently and uniformly randomized for each $(q, r_a, r_b)$ instance, so that the original ordering of $r_a$ and $r_b$ is hidden from the annotator.
Model identifiers, generation hyperparameters, and any other metadata that could leak the source system are stripped from the displayed text.
The interface exposes the following fields:

\begin{DefinitionBox}
\textbf{Query.}\quad The original DeepResearch-style query $q$.

\textbf{Report~1 / Report~2.}\quad The two anonymized candidate reports, each rendered with markdown formatting and inline citations preserved.

\textbf{Preference.}\quad A radio-button forced choice between ``Prefer Report~1'' and ``Prefer Report~2''. Ties are not allowed at the per-annotator level; the redundancy ($\geq 3$ annotators per pair, majority vote) absorbs residual ambiguity at the dataset level.

\textbf{Dominant Reason.}\quad A required short text field capturing the main factor behind the decision (e.g., ``Report~2 covers two more requested sub-questions and Report~1 contradicts itself on the timeline'').

\textbf{Confidence.}\quad An optional 3-point self-reported confidence (low / medium / high), used only to flag low-confidence pairs for re-annotation.
\end{DefinitionBox}

\textbf{Reading Procedure.}\quad
To prevent the common failure mode in which annotators commit to a preference after only skimming the first few paragraphs, we enforce the following reading order:
(1) Read the query and explicitly note, in scratch space, the information needs it expresses (sub-questions, requested scope, intended audience or tone, output structure, etc.);
(2) Read Report~1 in full, then Report~2 in full, refraining from cross-comparison during the first pass;
(3) Re-read both reports side-by-side, comparing them on each of the four evaluation dimensions defined below;
(4) Only then commit to a preference and write the dominant-reason justification.
We encourage annotators to spend on average $10$--$15$ minutes per pair; pairs annotated in under $3$ minutes are flagged and resampled.

\textbf{Evaluation Dimensions.}\quad
Annotators consider the following four dimensions, in keeping with the description in~\Cref{sec:pre_dataset}.
We deliberately do \emph{not} prescribe fixed weights, because part of the value of the resulting preference data is that it captures how humans implicitly trade these dimensions off across heterogeneous queries.
For each dimension we list both positive and negative signals to guide the comparison.

\begin{itemize}
\setlength{\itemsep}{2pt}
\setlength{\parsep}{0pt}
\setlength{\parskip}{0pt}
\item \textbf{Alignment with Information Need.}\quad
Does the report answer the question that was actually asked, at the requested granularity, scope, and output format?
Positive signals: addressing all sub-questions implied by the query; staying on topic; matching any explicitly requested structure (e.g., ``compare X and Y'', ``list five examples'').
Negative signals: drift to tangential topics; answering a different (often easier) question; ignoring an explicit constraint in $q$.

\item \textbf{Usefulness.}\quad
Conditional on alignment, how informative and decision-supporting is the report for a reader with the information need expressed in $q$?
Positive signals: concrete facts, numbers, named entities, evidence-grounded claims, and citations a reader can verify; clear takeaways.
Negative signals: vague generalities, padding, hedging without substance, or mechanical restatement of the query.

\item \textbf{Completeness.}\quad
Does the report cover the relevant facets of the topic at appropriate depth, without notable omissions of evidence that a knowledgeable reader would expect?
Positive signals: balanced coverage of multiple sub-aspects; mention of important counter-evidence or alternative views when relevant; appropriate depth on the central question.
Negative signals: one-sided treatment; missing whole sub-questions; superficial enumeration without analysis.

\item \textbf{Coherence.}\quad
Is the report well-organized and internally consistent at both global and local levels?
Positive signals: clear logical flow; section structure that mirrors the analytical narrative; consistent terminology and timeline; smooth transitions.
Negative signals: contradictions between sections; abrupt topic shifts; repeated content; confusing reference resolution; broken citation order.
\end{itemize}

\textbf{Decision Rule and Forced Choice.}\quad
Annotators make a holistic judgement rather than a strict weighted sum.
When the four dimensions disagree---for example, when one report is more aligned but less complete---we ask annotators to use the priority order \emph{Alignment} $\succ$ \emph{Usefulness} $\succ$ \emph{Completeness} $\succ$ \emph{Coherence} as a tie-breaker, on the grounds that a fluent and well-structured report that does not actually answer $q$ is of limited use in DeepResearch.
The forced binary choice is intentional: we observed in pilot studies that allowing per-annotator ties effectively concentrates probability mass on borderline pairs, which then becomes the noisiest part of the dataset.
Residual ambiguity is instead handled at the aggregation stage by majority voting over $\geq 3$ independent annotations; pairs that fail to reach a majority after a fourth annotator is added are discarded rather than coerced into a label.

\textbf{Biases to Avoid.}\quad
We explicitly warn annotators against, and our protocol structurally controls for, the following biases that are known to corrupt preference data for long-form generation:

\begin{itemize}
\setlength{\itemsep}{2pt}
\setlength{\parsep}{0pt}
\setlength{\parskip}{0pt}
\item \emph{Length bias}: longer reports are not automatically better. Annotators are reminded that padding, redundancy, and over-broad coverage outside the scope of $q$ should count against a report.
\item \emph{Surface formatting bias}: rich markdown headings, bold text, and bulletization should not by themselves drive preference; they only matter insofar as they improve coherence or alignment with $q$.
\item \emph{Citation-count bias}: more citations are not automatically better. Annotators check whether citations are relevant and whether they support the surrounding claim, rather than counting them.
\item \emph{Position bias}: mitigated at the system level by independently randomizing the Report~1 / Report~2 ordering for each pair, so that no annotator can develop a stable left/right preference.
\item \emph{Source/identity bias}: model identifiers and generation metadata are stripped from the rendered text, so annotators cannot recognize and prefer their favorite system.
\item \emph{Domain-familiarity bias}: queries are rotated across annotators, and an annotator may flag a pair as ``outside expertise'', in which case the pair is reassigned. This avoids over-weighting any single annotator's idiosyncratic taste in any one domain.
\item \emph{First-impression bias}: addressed by the mandatory full read of both reports before the dimension-by-dimension comparison.
\end{itemize}

\textbf{Quality Control.}\quad
We continuously monitor annotation quality through three mechanisms.
First, $5\%$ of every batch consists of \emph{seeded gold pairs} drawn from the calibration set; any annotator whose accuracy on gold pairs falls below $80\%$ across a sliding window of $50$ items is paused and re-trained.
Second, we compute the inter-annotator agreement (Fleiss' $\kappa$) within each batch and flag batches with $\kappa < 0.4$ for review.
Third, the lead annotator manually audits a random $2\%$ sample of finalized triples and the full set of low-confidence ones.
Pairs rejected during audit are returned to the pool for re-annotation by a fresh set of annotators.
After all filtering and aggregation, the released $\mathcal{D}$ contains only triples whose final labels are supported by a strict majority of $\geq 3$ qualified annotators and that have passed the audit step.

\textbf{Ethics, Compensation, and Working Conditions.}\quad
Annotators were recruited under a written agreement that explicitly describes the task, the data they will see, and the intended research use of the resulting dataset.
Compensation is set on a per-pair basis at a rate that, for the average expected reading time, exceeds the local minimum hourly wage; pairs flagged for re-annotation are paid in full to the original annotator.
To avoid fatigue effects, no annotator is allowed to annotate more than $30$ pairs per day or work for more than $4$ continuous hours without a break.
All annotators were informed that they could withdraw from the study at any time without penalty, and the rendered reports were screened for personally identifiable information before being shown.

\section{Metrics for Human Preference}\label{app:metrics}

We evaluate the performance of preference modeling using two complementary metrics.
Given a preference dataset
$\{(q_i, r_{\text{acc}}^{(i)}, r_{\text{rej}}^{(i)})\}_{i=1}^N$
with scalar scores $S(\cdot)$, we first report \emph{preference accuracy}, defined as
\begin{equation}
\mathrm{Pref. Acc.}
= \frac{1}{N}\sum_{i=1}^N
\mathbb{I}\!\left[
S\!\left(r_{\text{acc}}^{(i)}\right)
>
S\!\left(r_{\text{rej}}^{(i)}\right)
\right],
\end{equation}
which is equivalent to the area under the ROC curve (AUC) for pairwise preference judgments.
To further quantify the \emph{magnitude} and \emph{stability} of preference separation, we report the paired Cohen's $d$~\cite{diener2010cohen}.
Let $\Delta_i = S(r_{\text{acc}}^{(i)}) - S(r_{\text{rej}}^{(i)})$ denote the score difference for query $q_i$; the paired effect size is defined as
\begin{equation}
\text{Cohen's $d$}
= \frac{\mathbb{E}[\Delta]}{\sqrt{\mathrm{Var}(\Delta)}}.
\end{equation}
While preference accuracy (AUC) reflects ranking correctness, paired Cohen's $d$ captures the standardized strength of score separation at the query level, providing a complementary view of preference quality.

\section{Rollout Speed-up for MaMs workflow}\label{app:speed}


\begin{figure}[htp]
\centering
\includegraphics[width=0.7\linewidth]{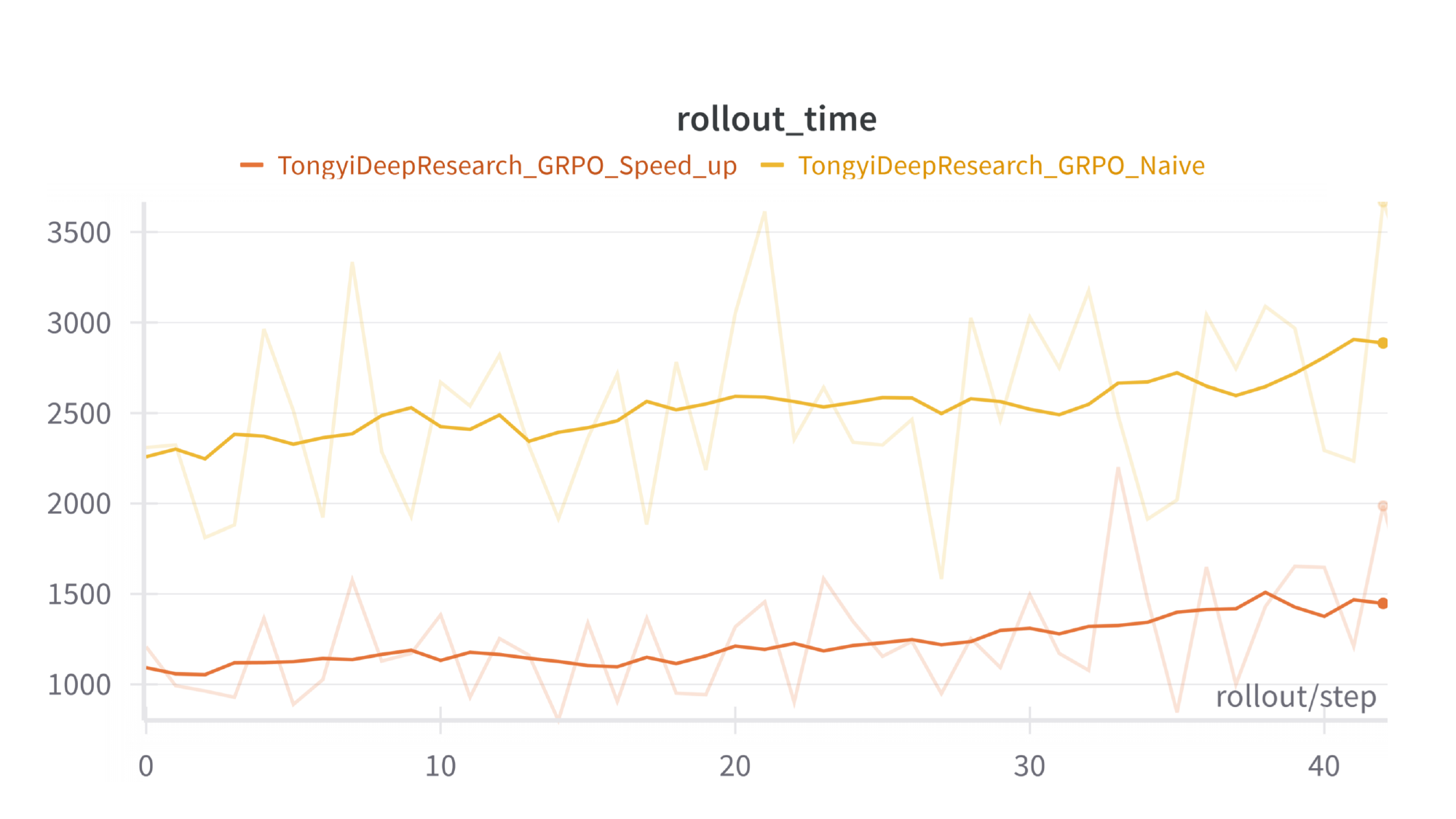}
\caption{Speed-up achieved by overlapping multiple micro-batches using the asynchronous event loop. The concurrency-limited scheduling allows high-latency API calls to run in parallel, maximizing resource utilization and reducing the effective runtime of the stage from linear in the dataset size $|\mathcal{D}|$ to approximately $|\mathcal{D}|/C$, where $C$ is the concurrency limit.}
\label{fig:speed_up}
\end{figure}

To address the latency bottlenecks inherent in sequential processing, specifically within I/O-bound Large Language Model (LLM) interactions, we introduce a parallel execution mechanism in the MaMs workflow.
The baseline implementation, referred to as the \textit{Naive Linear Pipeline}, processes the entire dataset $\mathcal{D}$ sequentially through a chain of agents.
In this mode, the total execution time $T_{naive}$ is the summation of the processing time for all samples, where network latency accumulates linearly.

To optimize efficiency, we developed the \textit{Linear Concurrent Pipeline}, which implements data parallelism via asynchronous micro-batching.
The pipeline divides the agent execution flow into three stages: pre-processing, concurrent execution, and post-processing. The acceleration focuses on the concurrent stage, where the input dataset $\mathcal{D}$ is partitioned into a sequence of micro-batches $B = \{b_1, b_2, \dots, b_m\}$, each with a configurable size $S_{micro}$.

We leverage an event loop to manage a pool of asynchronous tasks subject to a concurrency limit $C$. The scheduling algorithm operates as follows:
\begin{enumerate}
\item A sliding window maintains a set of active tasks $\mathcal{T}$, ensuring $|\mathcal{T}| \le C$.
\item As long as the active task slots are available ($|\mathcal{T}| < C$), new micro-batches are dequeued, and corresponding agent tasks are spawned immediately using \texttt{asyncio}.
\item Upon task completion, results are collected via callbacks into a synchronized queue, and the window slides forward to admit pending micro-batches.
\end{enumerate}

As shown in \Cref{fig:speed_up}, by overlapping the high-latency API calls across multiple micro-batches, the framework significantly maximizes resource utilization.
This approach effectively safeguards against blocking operations, reducing the theoretical time complexity for the concurrent stage from $O(|\mathcal{D}|)$ to approximately $O(|\mathcal{D}|/C)$, bounded primarily by external API rate limits rather than local execution speed.

\section{Preference Performance of Rubric Generator trained with GSPO}\label{app:gspo}

In this section, we will show the preference accuracy (AUC) and paired Cohen's $d$ of the rubric generator trained by GSPO in \Cref{tab:gspo}.

\begin{table}[htp]
\centering
\caption{Preference performance of rubric generators trained by GSPO.}
\label{tab:gspo}
\resizebox{0.76\linewidth}{!}{
\begin{tabular}{l|l|c|c}
\toprule \hline
\textbf{Model} & \textbf{Method} & \textbf{Pref. Acc./AUC (\%)} & \textbf{Paired Cohen's $d$}\\ \hline
\multirow{2}{*}{\textbf{Qwen3-30B-A3B}}  &  GRPO with Hybrid Reward & $65.68$ & $0.376$ \\ \cline{2-4}
  &  GSPO with Hybrid Reward & $62.02$ & $0.337$ \\ 
\hline \bottomrule
\end{tabular}
}
\end{table}

As discussed in \Cref{sec:entropy}, the rubric generator trained with GSPO exhibits substantially higher rollout entropy than its GRPO-trained counterpart, which is undesirable for our setting. 
Consequently, we adopt GRPO for training the rubric generator.

\section{Analysis on Entropy over Two RL Algorithms}\label{sec:entropy}

\begin{figure}[t]
\centering
\subfigure[]{
\includegraphics[width=0.4\linewidth]{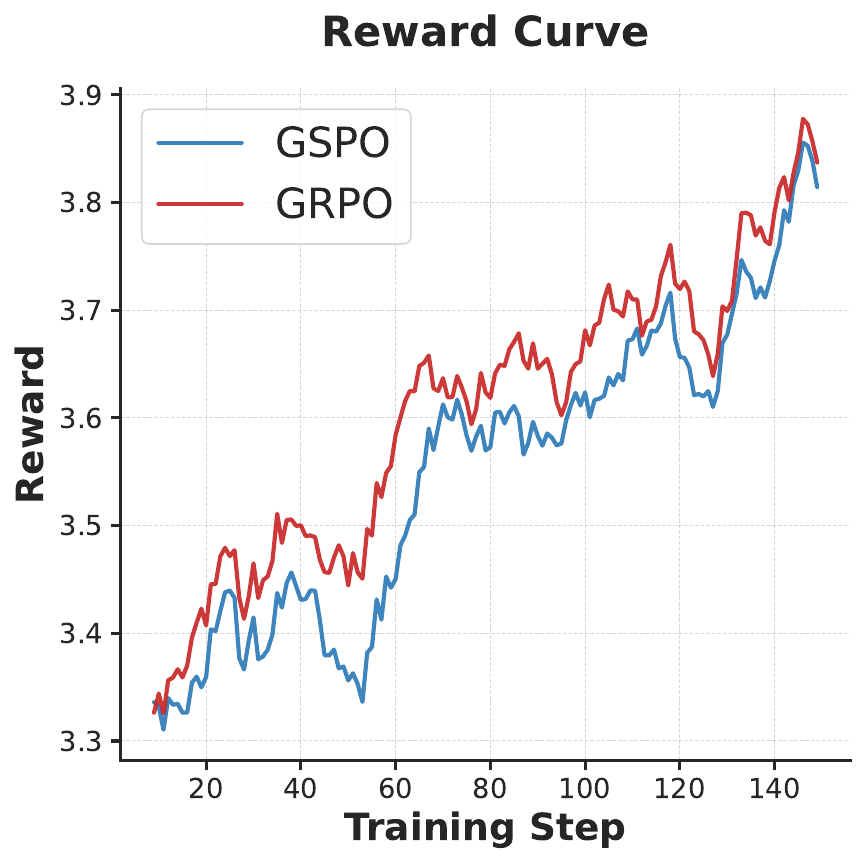}}
\subfigure[]{
\includegraphics[width=0.4\linewidth]{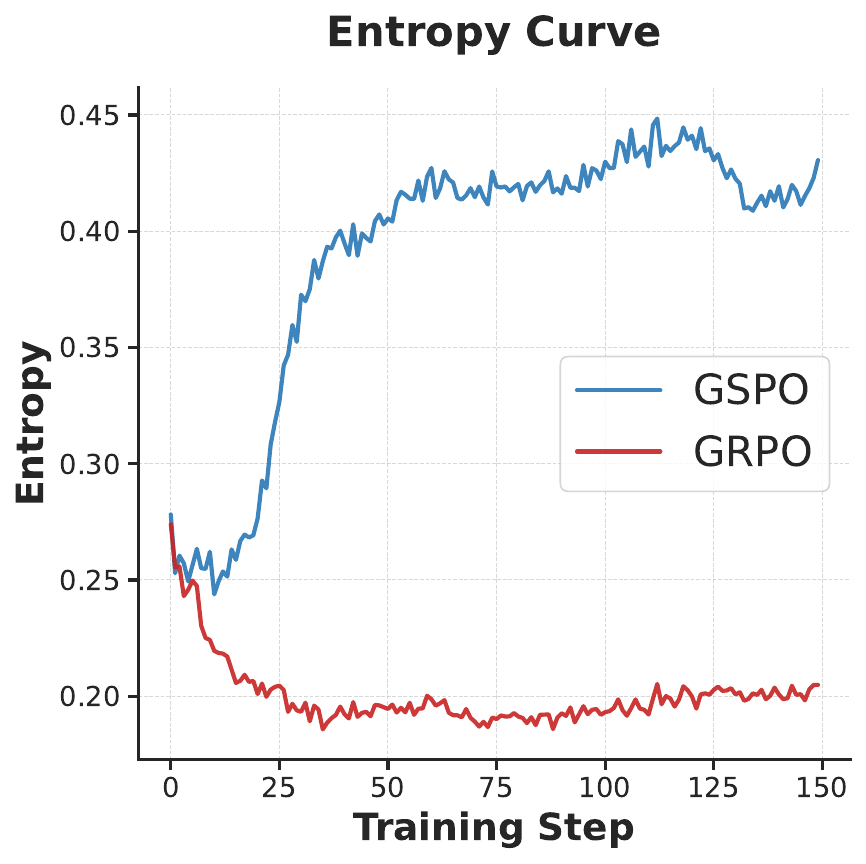}}
\caption{
Comparison between GSPO and GRPO during training rubric generators (Qwen3-30B-A3B).
(a) Reward curves of generated rollouts under the two algorithms, showing nearly identical reward values.
(b) Entropy of generated rollouts, where GSPO consistently exhibits higher entropy than GRPO.
}
\label{fig:analysis}
\vspace{-4mm}
\end{figure}

Since Qwen3-30B-A3B is a Mixture-of-Experts model, applying GRPO during training may introduce a mismatch between the expert routing used for optimization and that employed during rollout.
To mitigate this issue, we additionally explore GSPO~\cite{zheng2025group} for training the rubric generator.
Although GSPO and GRPO share identical training configurations, we observe that rubric generators trained with GSPO consistently produce rollouts with higher entropy, as shown in~\Cref{fig:analysis}, despite achieving nearly identical reward values on the generated samples.
We attribute this behavior to GSPO's sequence-level optimization scheme, where importance weighting and clipping are performed over entire responses rather than individual tokens.
This design reduces sensitivity to local token-level deviations and allows multiple realizations with similar global rewards to coexist, leading to increased output diversity.
In contrast, GRPO applies group-wise relative advantages over complete rollouts but relies on token-level likelihood ratios, which implicitly impose stronger structural constraints on the generated rubric format.
Given that rubric generation prioritizes stability, consistency, and preference alignment rather than linguistic diversity, we adopt GRPO as it better matches the mode-seeking nature of the task.
The performance of rubric generators trained with GSPO is reported in Appendix~\ref{app:gspo}.

\section{Case Study of Rubric List}
We show a case about the rubric list of a question in the following:
\begin{lstlisting}
{
  "question": "Please generate an analysis report on common network failures.",
  "rubrics": [
    {
      "title": "Coverage of Common Failures",
      "description": "Key criterion: The report must identify and describe multiple common types of network failures, such as DNS issues, IP address conflicts, or physical connection interruptions.",
      "weight": 5
    },
    {
      "title": "Inclusion of Core Analysis",
      "description": "Key criterion: The report must analyze each mentioned network failure rather than merely listing their names.",
      "weight": 5
    },
    {
      "title": "Clear Structure",
      "description": "Key criterion: The report should have a clear organizational structure, such as an introduction, categorized analysis of different failures, and a conclusion.",
      "weight": 5
    },
    {
      "title": "Analysis of Causes and Symptoms",
      "description": "Important criterion: The report should explain the typical symptoms and possible causes of each network failure, establishing a clear causal relationship.",
      "weight": 4
    },
    {
      "title": "Provision of Troubleshooting Methods",
      "description": "Important criterion: The report should provide concrete and actionable troubleshooting steps or solution suggestions for each type of failure.",
      "weight": 4
    },
    {
      "title": "Clear and Understandable Explanation",
      "description": "Important criterion: When explaining technical concepts (such as DNS or IP addresses), the report should strive to be clear and accurate so that non-expert readers can understand it.",
      "weight": 3
    },
    {
      "title": "Professional and Objective Tone",
      "description": "Important criterion: The report should maintain a professional and objective tone, avoiding overly colloquial or subjective expressions.",
      "weight": 3
    },
    {
      "title": "Systematic Classification of Failures",
      "description": "Optional criterion: The report may systematically categorize network failures based on their nature (e.g., hardware, software, configuration issues) to enhance clarity.",
      "weight": 2
    },
    {
      "title": "Inclusion of Preventive Measures",
      "description": "Optional criterion: The report may further propose preventive measures and best practices to avoid common network failures.",
      "weight": 2
    },
    {
      "title": "Use of Concrete Examples",
      "description": "Optional criterion: The report may use concrete scenarios or cases to illustrate failure phenomena and solutions, improving readability.",
      "weight": 1
    },
    {
      "title": "Technical Errors",
      "description": "Error criterion: The report provides incorrect technical explanations, causes, or solutions that may mislead readers.",
      "weight": -2
    },
    {
      "title": "Listing Without Analysis",
      "description": "Error criterion: The report merely lists failure names without providing any analysis of causes, symptoms, or solutions.",
      "weight": -2
    },
    {
      "title": "Inclusion of Irrelevant Information",
      "description": "Error criterion: The report includes content unrelated to common network failures, such as in-depth discussion of unrelated software programming errors.",
      "weight": -1
    }
  ],
  "topic": "Science & Technology",
  "rubric_count": 13
}
\end{lstlisting}

\section{Dataset and Rubric Statistics}\label{app:data_stats}

We report detailed statistics of the preference dataset $\mathcal{D}$ and the generated rubrics to provide additional insight into data quality and composition.

\textbf{Rubric Count Distribution.}\quad
Each rubric is generated as a structured list of evaluation criteria.
Across the train/valid/test splits ($5{,}651$ queries total), the rubric generator produces a mean of $13.3$ items per query (median $13$), with $94.3\%$ of rubrics containing $10$--$16$ items and the full range spanning $8$--$23$.
This convergence occurs naturally during RL training without explicit count constraints, indicating that the generator discovers a stable level of evaluation granularity.

\textbf{Rubric Weight Structure.}\quad
\Cref{tab:weight_dist} shows the distribution of rubric item weights across all $75{,}251$ items.
Positive-weight items ($78.1\%$) reward desirable report qualities at three tiers: critical (weight $5$), important ($3$--$4$), and optional ($1$--$2$).
Negative-weight items ($21.9\%$) penalize specific errors or omissions (weight $-1$ to $-2$).
The generator thus learns to produce rubrics with both reward and penalty criteria, mirroring the structure of expert-designed rubrics~\cite{gunjal2025rubrics} without explicit supervision of this property.

\begin{table}[htp]
\centering
\small
\caption{Distribution of rubric item weights across all splits.}
\label{tab:weight_dist}
\vspace{-2mm}
\begin{tabular}{crrrl}
\toprule
\textbf{Weight} & \textbf{Count} & \textbf{\%} & & \textbf{Role} \\
\midrule
$5$ & 19{,}214 & 25.5 & \multirow{4}{*}{\rotatebox[origin=c]{90}{\small reward}} & Critical \\
$4$ & 13{,}895 & 18.5 & & Important \\
$3$ & 12{,}644 & 16.8 & & Important \\
$2$ & 6{,}488 & 8.6 & & Optional \\
$1$ & 6{,}529 & 8.7 & & Optional \\
\midrule
$-1$ & 3{,}141 & 4.2 & \multirow{2}{*}{\rotatebox[origin=c]{90}{\small pen.}} & Minor error \\
$-2$ & 13{,}340 & 17.7 & & Major error \\
\bottomrule
\end{tabular}
\end{table}

\section{Future work}\label{app:limit}

Several directions offer promising opportunities for future research.  
First, the preference formulation could be extended beyond pairwise comparisons to leverage richer preference signals, such as rankings or graded scores, enabling more fine-grained learning of human preferences.
Second, future work could focus on improving the assessment of novelty, creativity, factuality, and reasoning depth, for example, by combining more sophisticated LLM evaluations with targeted human feedback to reduce subjectivity and increase reliability.
Third, developing more principled approaches to reduce dependence on LLM-based evaluation, potentially through self-consistency checks or hybrid human-LLM validation, could enhance the stability and interpretability of the training process.
Finally, broader evaluations across scholarly search, domain-specific corpora, and alternative report formats would better characterize the transferability of preference-trained rubric generators.

\end{document}